\documentclass[amsthm]{statsoc}

\usepackage{geometry}
\usepackage{paralist} 

\makeatletter
\renewcommand\paragraph{\@startsection{paragraph}{4}{0pt}%
	{0pt}
	{-\parindent}
	{\bfseries}}
\makeatother

\usepackage{graphicx}
\usepackage[textwidth=8em,textsize=small]{todonotes}
\usepackage{amsmath}
\usepackage{natbib}

\usepackage{amsfonts}
\usepackage{placeins}
\usepackage{makecell}
\usepackage{booktabs}

\usepackage{bm}
\usepackage{tikz}
\usetikzlibrary{bayesnet}
\usepackage{xspace}
\usepackage{amssymb}
\usepackage{float}
\usepackage{multirow}

\usetikzlibrary{automata,arrows}
\usepackage[hyphens]{url}
\usepackage[hidelinks]{hyperref}
\hypersetup{breaklinks=true}
\newcommand{\ie}{i.e.\xspace}
\newcommand{\eg}{e.g.\xspace}

\newcommand{\eq}{Eq.\xspace}
\newcommand{\eqs}{Eqs.\xspace}
\newcommand{\fig}{Fig.\xspace}

\newcommand{\acro}[1]{\textsc{#1}\xspace}
\newcommand{\mcmc}{\acro{mcmc}}
\newcommand{\vi}{\acro{vi}}

\newcommand{\pdf}{\acro{pdf}}
\newcommand{\termkl}{\acro{kl}}

\newcommand{\auc}{\acro{auc}}
\newcommand{\NRMSE}{\acro{nrmse}}
\newcommand{\mse}{\acro{mse}}
\newcommand{\Rsqd}{\text{R}^2}
\newcommand{\elbo}{\acro{elbo}}

\newcommand{\our}{\acro{bsim}}

\newcommand{\comment}[1]{}

\newcommand{\tikzcircle}[2][circle,fill=white,draw=black]{\tikz[baseline=-0.5ex]\draw[#1,radius=#2] (0,0) circle ;}%
\newcommand{\tikzcirclefild}[2][circle,fill=gray!25,draw=black]{\tikz[baseline=-0.5ex]\draw[#1,radius=#2] (0,0) circle ;}%
\newcommand{\tikzcircleblack}[2][circle,fill=black,draw=black]{\tikz[baseline=-0.5ex]\draw[#1,radius=#2] (0,0) circle ;}%

\tikzstyle{tikprior} = [circle, fill=black,minimum size=8pt,radius=2pt, inner sep=0pt, node distance=0.4]

\newcommand{\mat}[1]{\mathbf{#1}}
\renewcommand{\vec}[1]{ \mathbf{#1} } 

\newcommand{\vect}[1]{\boldsymbol{\mathbf{#1}}}

\newcommand{\feat}{\vect{\phi}}
\newcommand{\indfeat}{\vect{v}}
\newcommand{\capitalfeat}{\vect{\Phi}}
\newcommand{\capitalindfeat}{\vect{V}}
\newcommand{\storelocation}{\vec{l}}
\newcommand{\indlocation}{\vec{m}}
\newcommand{\indrev}{r_{n}} 	
\newcommand{\indrevstore}{r_{ns}} 	
\newcommand{\storerev}{r_{s}} 	
\newcommand{\storecovariance}{\vect{\Sigma}}
\newcommand{\storemean}{\vect{\mu}}
\newcommand{\lambdapar}{\vect{\lambda}}
\newcommand{\errorterm}{\sigma^2}
\newcommand{\storesigma}{\sigma^2_s}
\newcommand{\storeserror}{\varepsilon}
\newcommand{\maxdist}{d_T}
\newcommand{\distns}{d_{ns}}

\newcommand{\Rev}{\mathbf{Y}}
\newcommand{\predRev}{\mathbf{\hat{Y}}}
\newcommand{\vecbeta}{\vect{\beta}}
\newcommand{\veclambda}{\vec{\lambda }}

\newcommand{\dataset}{\mathcal{D}}
\newcommand{\x}{\vec{x}}

\newcommand{\vectheta}{\vec{\Theta}}

\newcommand{\pns}{p_{ns}}

\newcommand{\imat}{\mat{I}}

\newcommand{\region}{\tau}

\usepackage{etoolbox}

\makeatletter
\patchcmd{\@makecaption}
{\parbox}
{\advance\@tempdima-\fontdimen2} 
{}{}
\makeatother  
\usepackage[caption = false]{subfig}

\title[A Variational Bayesian Spatial Interaction Model]{A variational Bayesian spatial interaction model for estimating revenue and demand at business facilities}      
\author[Shanaka Perera {\it et al.}]{Shanaka Perera}
\address{University of Warwick, Coventry, United Kingdom}
\email{s.perera@warwick.ac.uk}
\coaddress{Shanaka Perera, Department of Computer Science, University of Warwick, CV4 7AL, UK.}
\author[]{Virginia Aglietti}
\address{University of Warwick, Coventry, United Kingdom}
\address{The Alan Turing Institute, London, United Kingdom} 
\author[]{Theodoros Damoulas}
\address{University of Warwick, Coventry, United Kingdom}
\address{The Alan Turing Institute, London, United Kingdom} 

\begin{document}
	
\begin{abstract}
We study the problem of estimating potential revenue or demand at business facilities and understanding its generating mechanism. This problem arises in different fields such as operation research or urban science, and more generally, it is crucial for businesses' planning and decision making. We develop a Bayesian spatial interaction model, henceforth \our, which provides probabilistic predictions about revenues generated by a particular business location provided their features and the potential customers' characteristics in a given region. \our explicitly accounts for the competition among the competitive facilities through a probability value determined by evaluating a store-specific Gaussian distribution at a given customer location. We propose a scalable variational inference framework that, while being significantly faster than competing Markov Chain Monte Carlo inference schemes, exhibits comparable performances in terms of parameters identification and uncertainty quantification. We demonstrate the benefits of \our in various synthetic settings characterised by an increasing number of stores and customers. Finally, we construct a real-world, large spatial dataset for pub activities in London, UK, which includes over 1,500 pubs and 150,000 customer regions. We demonstrate how \our outperforms competing approaches on this large dataset in terms of prediction performances while providing results that are both interpretable and consistent with related indicators observed for the London region.   
\end{abstract}
	
	
\section{Introduction}
\label{section:intro}
Understanding the interaction between business facilities and consumer preferences is a prime factor of success for industries such as retail, healthcare and hospitality. Therefore, accurate predictions of potential sales at business locations are becoming crucial for planning and decision-making in the current ecosystem. Indeed, the continuous growth in e-commerce \citep{retailSalesData} is threatening the existence of traditional retail stores. We propose a Bayesian statistical methodology that, by capturing the relationship between attractiveness of the facility, distance between a business location and its customers, and demand in terms of buying power, allows to make probabilistic forecasts about potential revenue at a business facility while quantifying the uncertainty in these estimates. 

One of the earliest statistical models of customer behaviors when choosing shopping facilities is a spatial interaction model, called Law of Retail Gravitation \citep{reilly1929methods}, which was inspired by the Newtonian gravity model and formulated a customer's choice between two facilities as a function of their attractiveness and distances. \citet{Huff1963} subsequently extended this model to consider multiple facilities while providing a probabilistic interpretation for the spatial interactions between customers and facilities. 
In the following years, the Huff model \citep{Huff1963} was improved by replacing the single attractiveness term determined by floorspace with a composite index of a set of attributes at the facility, including economic and structural factors \citep{nakanishi1974parameter, DeGiovanni2014}. 
Most of the literature estimates the parameters of spatial interaction models by resorting to regression methods \citep{nakanishi1974parameter,fotheringham1980spatial, li2012assessing, bekti2018multiplicative} or by maximising the entropy with respect to some constraints \citep{fotheringham1983new,wilson2010entropy}. More recently, computationally intensive Markov Chain Monte Carlo (\mcmc) schemes have been proposed as an alternative inference method within the Bayesian framework for modelling origin-destination flows but do not offer capabilities in estimating total revenue or demand generated at the destination \citep{ellam2018stochastic,congdon2010random, lesage2010spatial}.  

Inspired by the literature on gravity models, we develop a Bayesian spatial interaction model, henceforth named \our, which provides probabilistic predictions about revenues generated at business facilities given their features and the potential customers' characteristics in a specified region in space. We model the probability of a customer visiting each facility in a region through Gaussian densities in geographic space. Specifically, each density is centered on a facility with variance that is further determined by its attractiveness which in turn modelled as a function of internal and external characteristics (\eg floorspace,  distance to public transport access points) and customer perspective (\eg customer rating). The revenues for each facility are then obtained by combining the probability of a customer visit with a proxy of the individuals buying power, which we assume to be a function of their socio-demographic characteristics. 
We adopt a Bayesian approach that enables us to adequately account for the uncertainty associated with the customer interactions with the facilities. Our framework not only gives accurate predictions but produces interpretable results that can support experts' decision-making processes. Moreover, this approach allows us to infer quantities at the business facility or customer level, such as revenue flow from customers to businesses. In \our, the posterior distributions of interest are intractable, and their approximation poses significant computational challenges. We address this issue by resorting to variational inference while also comparing with \mcmc approximation.  We demonstrate how our variational scheme is significantly faster compared to \mcmc used in the literature while providing comparable results in terms of parameter identification and uncertainty quantification.

In the literature, experiments on spatial interaction modelling are limited to small synthetic datasets or real-world aggregated data since acquiring granular level real-world data is usually expensive  \citep{berman2002locating,  aboolian2007competitive}. To address these constraints, we create a dataset that includes variables observed at a granular level for public houses (pubs) and customers. This is performed by combining large geospatial and non-geospatial data using open and commercial data sources. Additionally, we gather customer reviews from  Google's customer rating API, which covers a broader audience compared to the traditional survey methods found in the literature \citep{drezner2006derived}. We demonstrate the benefits of the proposed methodology on this real-world large scale dataset and show how \our outperforms competing approaches in terms of prediction performances. Furthermore, we illustrate how \our provides interpretable results consistent with other industry-related indicators observed for London.

Our main contributions are: \begin{inparaenum}[(a)] \item we develop a Bayesian spatial interaction model (\our) that can be used to make probabilistic predictions of revenues or demand generated at business facilities and formulates the relationship between distance and attractiveness of facilities jointly, using a facility-specific probability distribution; \item we propose a scalable variational inference and demonstrate its benefits compared to MCMC methods in a variety of experimental settings; \item we construct an unprecedented real-world large spatial dataset for pub activities at the most granular level along with customer characteristics at the postcode level, collated from multiple sources; and \item We show that our method provides the best predictive performance compared to competing approaches while providing inference at the level of customers and business facilities, delivering invaluable insights for planning and decision making. To the best of our knowledge, we are the first to demonstrate an application of a  Bayesian spatial interactions model on a large scale real-world dataset describing pub activities in the Greater London area with more than 1,500 business locations and 150,000 customer regions\end{inparaenum}.

This paper is organised as follows. In Section 2, we introduce \our and the related inference scheme. Then, we evaluate the model performance using synthetic experiments in Section 3. In Section 4, we introduce a comprehensive spatial database. Next, in Section 5, using the new dataset, we demonstrate the benefits of our approach by inferring the model parameters for a real-world case study. Finally, conclusions and future research directions are discussed in Section 6.

\section{Methodology}
\label{section:methodology}
We consider a regression problem for a given dataset $\dataset=\{\x_s,y_s\}_{s=1}^S$, where $\x_s \in \mathbb{R}^D$ represents the $s$-th store\footnote{We present the rest of the model in relation to the specific instantiation where a business location is a store, but this can be extended to other business facilities.} features and $y_s \in \mathbb{R}$ gives the revenue for the $s$-th store in a bounded region $\region$. Each feature vector $\x_s^\top = [\storelocation_s^\top, \feat_s^\top]$ includes the store location, which we denote by $\storelocation_s \in \mathbb{R}^2$, and additional store characteristics denoted by $\feat_s  \in \mathbb{R}^{D-2}$, \eg floor size. For notational convenience we will denote $S\times(D-2)$ matrix of all stores characteristics by $\capitalfeat$. We assume the existence of $N$ customers within $\region$ where $\indfeat_n$ is the $n$-th row of $\capitalindfeat \in \mathbb{R}^{N\times P}$ and represents the features of the $n$-th customer. $\indfeat_n$ includes the customer location, which we denote by $\indlocation_n \in \mathbb{R}^2$ and its characteristics such as income level. \\[-1 cm]

\subsection{Model Formulation}
\begin{figure}[h!]
\centering
	\begin{minipage}{.4\textwidth}
	\centering
	\includegraphics[width=6cm]{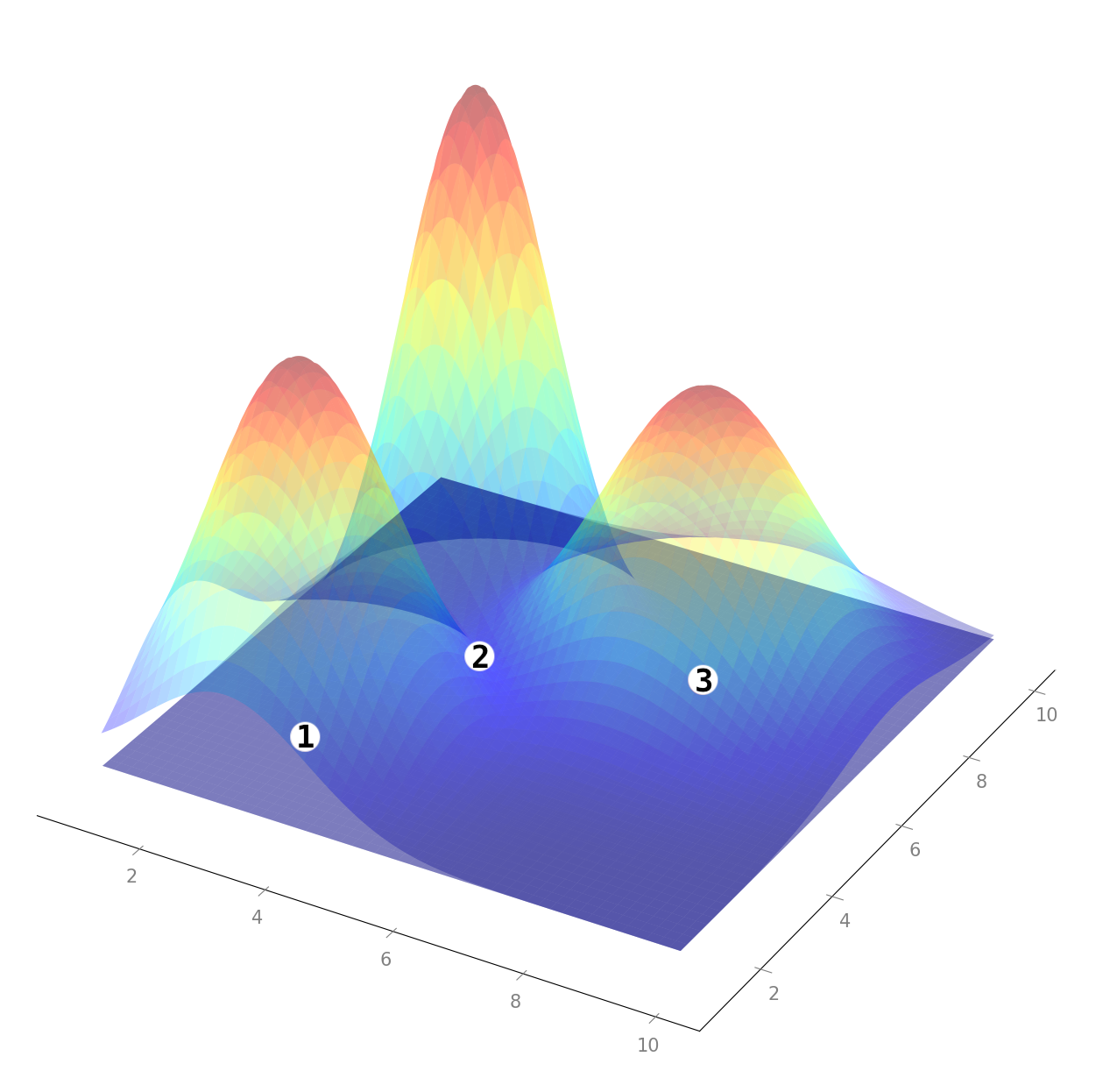}\\		
	\scriptsize (a) 
	\end{minipage}%
	\begin{minipage}{.4\textwidth}
	\centering
	\includegraphics[width=6cm]{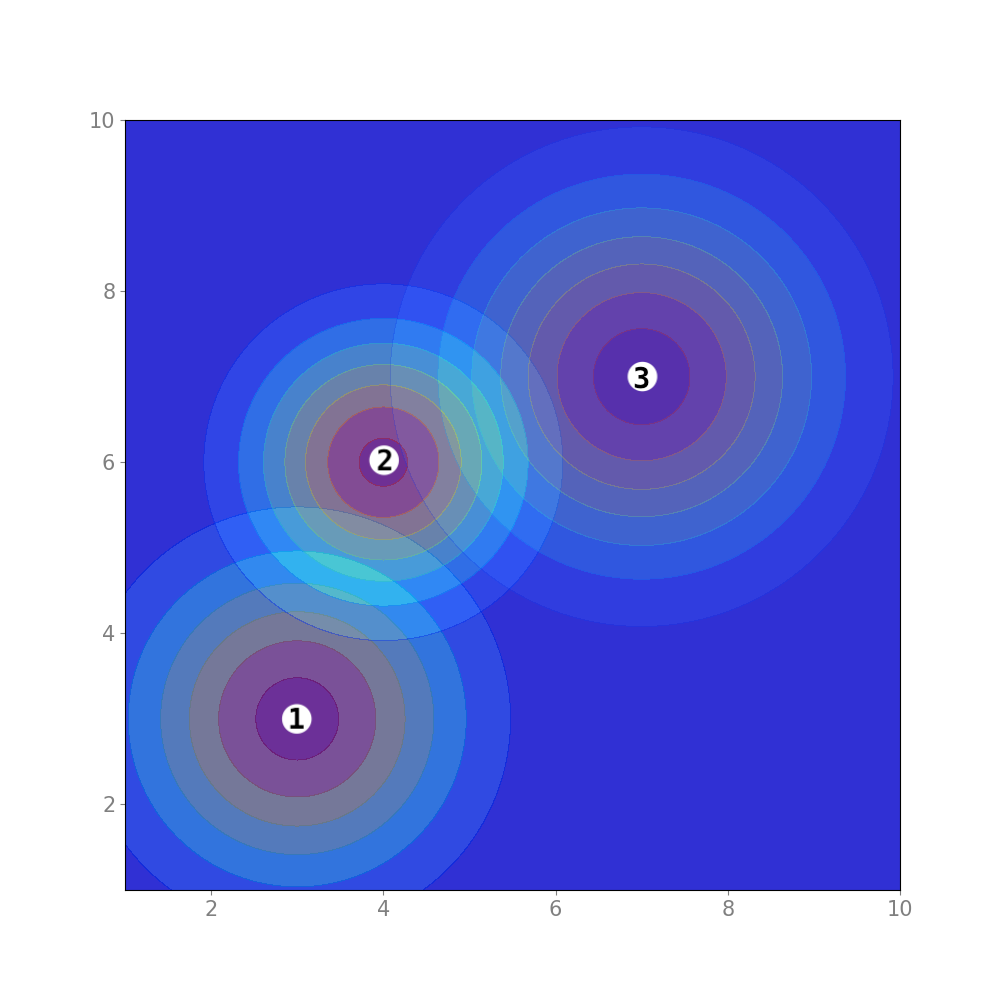}\\
	\scriptsize	(b) 
	\end{minipage}
\caption{Illustration of the \pdf of the Gaussian distribution centered on three sample Stores : (a) 3D visualisation; (b) 2D visualisation. The white dots indicate the store location and the numbers are used to identify the respective stores on 3D and 2D visualisations.}
\label{fig:Gaussians}
\end{figure}

The proposed Bayesian Spatial Interaction Model (\our) is characterised by $S$ Gaussian distributions, one for each store, which are uncorrelated a priori. Each Gaussian distribution, henceforth $\mat{Z}_s \sim \mathcal{N}(\storemean_s, \storecovariance_s)$, is centered on a store's location $\storemean_s = \storelocation_s$ and has a diagonal covariance matrix $\storecovariance_s = \storesigma\imat$. The variance $\storesigma$ captures level of ``attraction" of a customer to a store. We propose two different alternative models for variance. In the first model $\storesigma$ is written as a function of store specific coefficient  $\upsilon_s \in \mathbb{R}$ that is: 
\begin{align}
\label{eq:model1}
\storesigma = \exp ( \upsilon_s ),
\end{align}
In the second, we improve the specifications by denoting $\upsilon_s$ as a function of store characteristics:
\begin{align}
\label{eq:model2}
\upsilon_s = \lambdapar^\top \feat_s + \storeserror_s, 
\end{align}
where $\lambdapar \in \mathbb{R}^{D-2}$ represents a shared coefficients across the stores and $\varepsilon_s$ denotes the unobservable store characteristics. Evaluating the probability density function (\pdf) of the variable $\mat{Z}_s$ at $\indlocation_n$, which we denote by $\mat{Z}_s(\indlocation_n)$, allows us to capture the likelihood for the $n$-th customer to visit the $s$-th store based on their distance and on the store characteristics. For illustration purposes, consider three stores where each has a Gaussian distribution centred on the store, as shown in \fig~\ref{fig:Gaussians}.

Irrespective of the store's attractiveness, customer behaviour is not affected after a certain maximum distance to the store, known as ``consideration set'' in marketing. Therefore, we truncate the Gaussian distributions in \our and force their densities to be zero beyond a given distance $\maxdist$ from the store location. The truncated Gaussian \pdf is given by:
\begin{align}
\mat{Z}_s(\indlocation_n)  = \begin{cases}
\displaystyle\frac{\exp{\left(-\distns^2/2 \sigma_s^2\right)}}{2\pi \sigma_s^2 \left(1 - \exp{(-\maxdist^2/2\sigma_s^2)}\right)}, & 0 \leq \distns\leq \maxdist,\\
0, & \text{otherwise},
\end{cases}
\end{align}
where $\distns$ denotes the Euclidean distance between the store and customer  $\distns= ||\indlocation_n-\storelocation_s||_2 $;  see Appendix A for details. \fig~\ref{fig:TruncGaussians} demonstrates the truncated Gaussian densities corresponding to the distributions shown in  \fig~\ref{fig:Gaussians}.

\begin{figure}[h!]
\centering
 	\begin{minipage}{.4\textwidth}
	\centering
	\includegraphics[width=6cm]{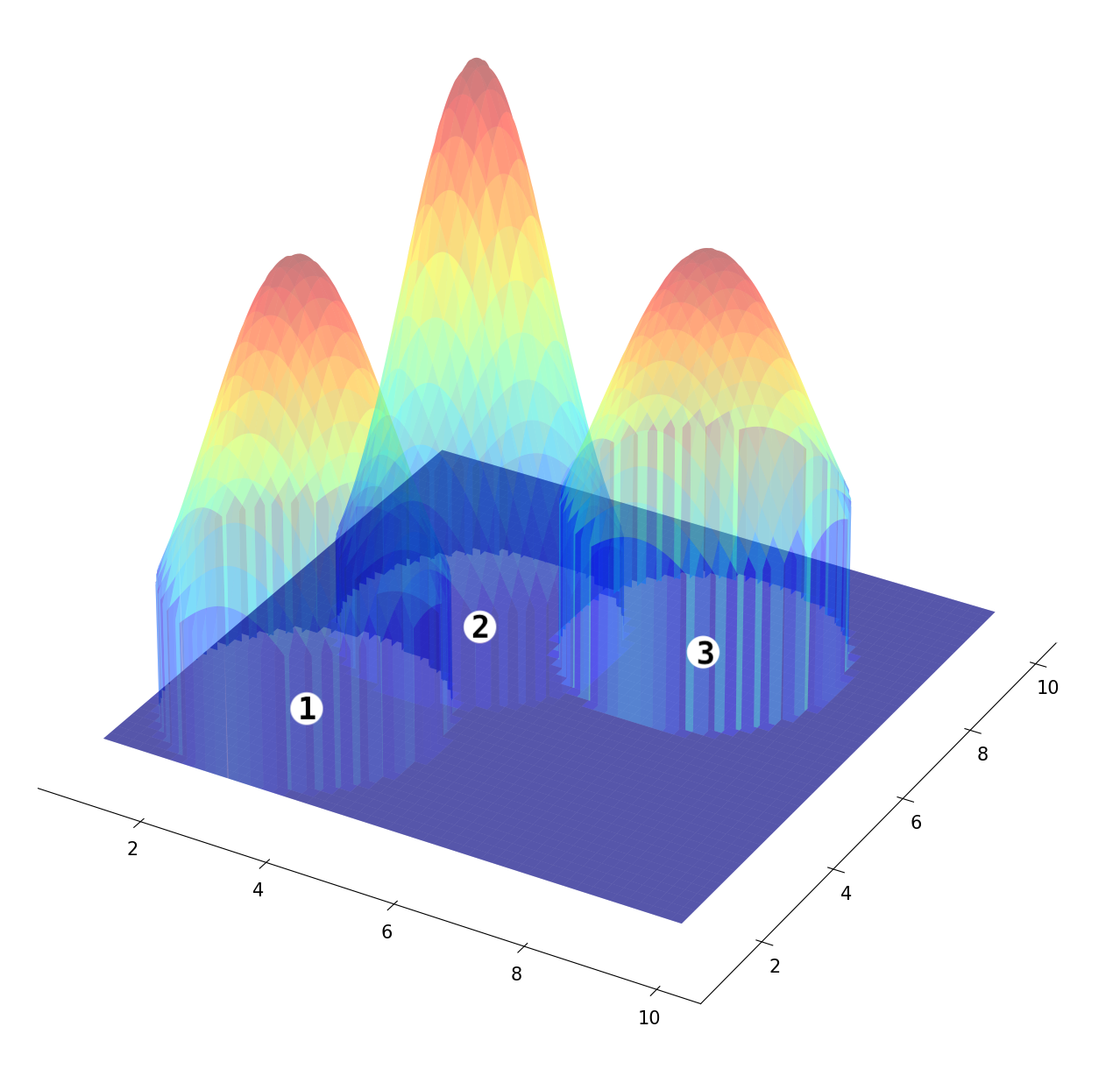}\\		
	\scriptsize (a) 
	\end{minipage}%
	\begin{minipage}{.4\textwidth}
	\centering
	\includegraphics[width=6cm]{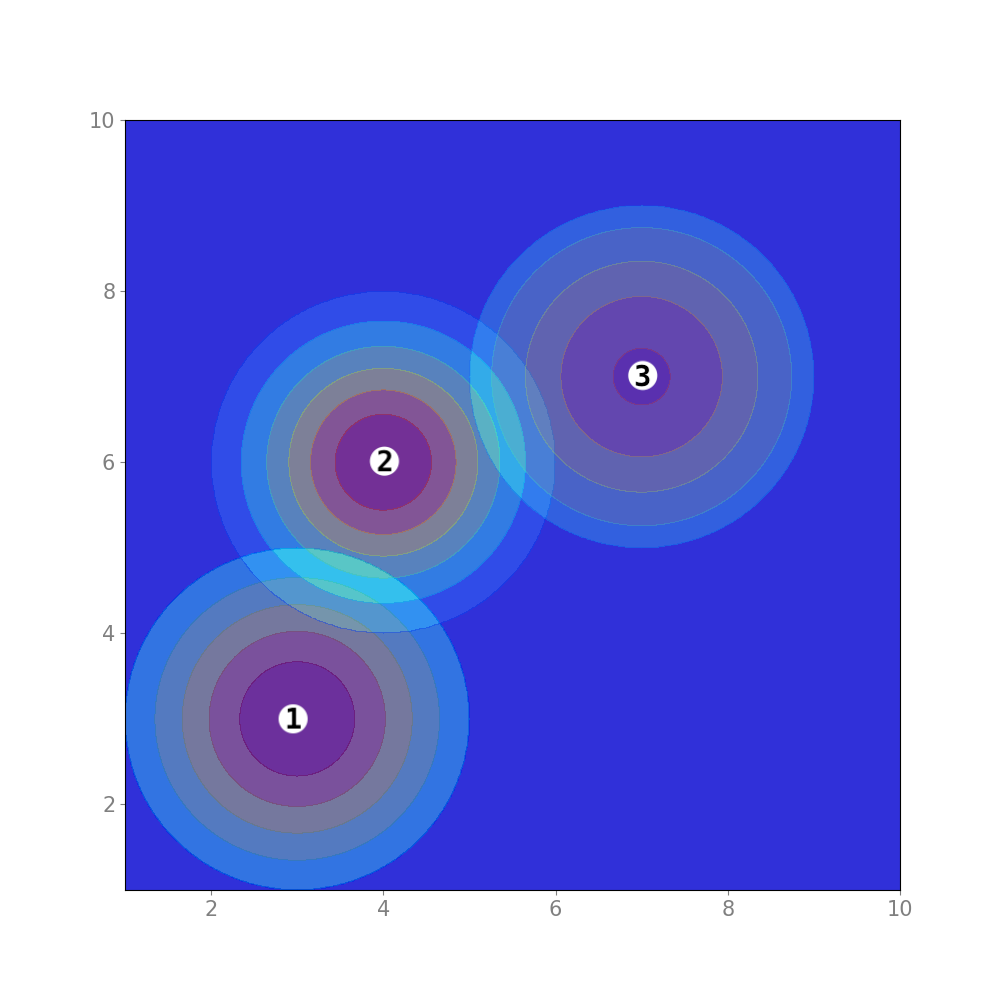}\\
	\scriptsize	(b) 
	\end{minipage}
\caption{Illustration of the Truncated Gaussian centered on three sample Stores: (a) 3D visualisation; (b) 2D visualisation. The white dots indicate the store location. There is a hard border around the distributions beyond which the \pdf is equal to zero.}
\label{fig:TruncGaussians}
\end{figure}

Given the truncated Gaussian distributions, we define the probability $\pns$ of a customer visiting the $s$-th store as: 
\begin{align}
\label{eq:responsability}
\pns = \frac{\mat{Z}_s(\indlocation_n)}{\sum_{j=1}^{S} \mat{Z}_j(\indlocation_n)}.
\end{align}

\begin{figure}[ht] 
\centering
	\begin{minipage}{\textwidth}
	\centering
	\includegraphics[width=13cm]{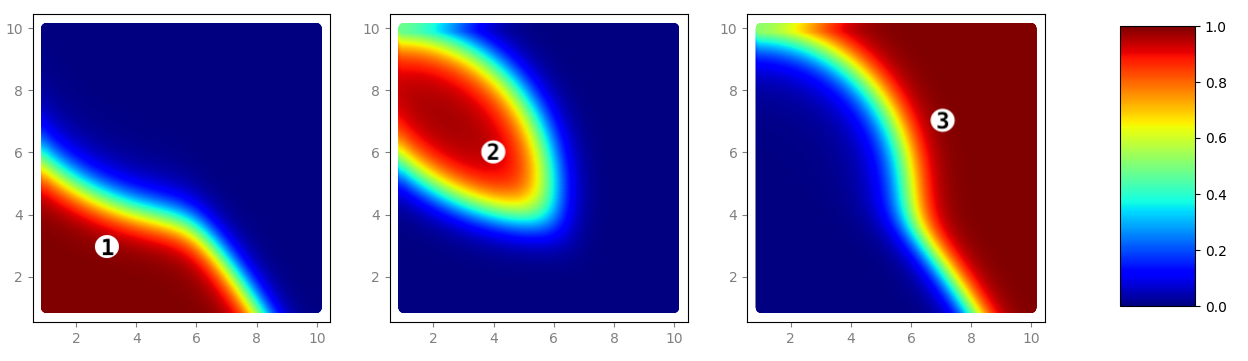}\\
	\scriptsize	(a) 
	\end{minipage}
	\begin{minipage}{\textwidth}
	\centering
	\includegraphics[width=13.1cm]{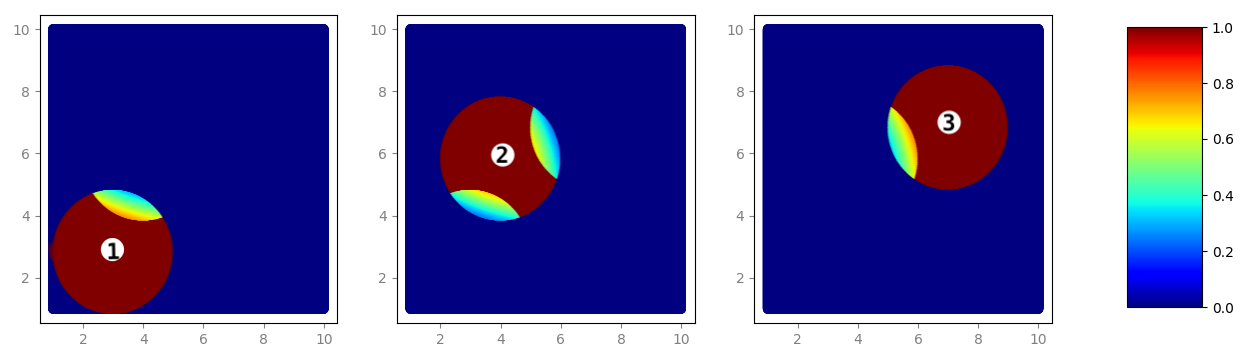}\\
	\scriptsize (b) 
	\end{minipage}
\caption{Illustration of the probability of customers visiting a store $\pns$: (a)  with none truncated Gaussian distribution; (b) with truncated Gaussian distribution. This is an indication of the competition in the area. The white dots indicate the store location, and the numbers are used to identify the respective stores on (a) and (b) plots.} 
\label{fig:p_ns}
\end{figure}

Note that we normalize the \pdf calculated for the customer with respect to the store by the total \pdf respect to all the stores within the consideration set to arrive at a value which falls in the interval of [0, 1]. Thus we assume that every customer chooses at least one store in their consideration set, but this can be relaxed by adding pseudo stores to account for unsatisfied demand or unobserved data. The value of $\pns$ capture the level of competition in the region $\region$ for a specific type of store. For instance, $\pns$ will be lower in competitive markets or areas while it will take higher values in non-competitive settings. This is illustrated in \fig~\ref{fig:p_ns} with respect to the non-truncated and truncated Gaussian distributions. 

The consumption function in economics determines the relationship between consumer spending and the various factors \citep{modigliani1954utility}. 
To model the amount budgeted by each customer for spending we propose a linear function $f(\cdot)$ which takes input $\indfeat_{n}[-\indlocation_n]$ representing the $P-2$ customer features obtained by discarding the location coordinates:
\begin{align}
\indrev = f(\indfeat_{n}[-\indlocation_n]) = \vecbeta^T\indfeat_{n}[-\indlocation_n],
\end{align}
where $\vecbeta \in \mathbb{R}^{P-2}$. This leads to the conventional Spatial interaction system \citep{Huff1963,Wilson1971,ellam2018stochastic}. Thus expenditure flow from customer $n$ to store $s$:
\begin{align}
\indrevstore = \indrev \times\pns,
\end{align}
where the amount each customer budgeted to spend $\indrev$ is weighed by the probability to visit the $s$-th store. The total revenue for the $s$-th store is:
\begin{align}
\storerev = \sum_{n=1}^{N} \indrev \pns = \sum_{n=1}^{N} \vecbeta^T\indfeat_{n}[-\indlocation_n] \frac{\mat{Z}_s(\indlocation_n)}{\sum_{j=1}^{S} \mat{Z}_j(\indlocation_n)}.
\end{align}

Henceforth we derive the model for the case where the store variance is a function of its features (\eq~\eqref{eq:model2}), since the limiting case where the store variance is store specific coefficient (\eq~\eqref{eq:model1}) is a trivial extension by setting $\lambda$ to zero.\\
\paragraph{Likelihood function:}
The likelihood of the observed stores' revenue  $\mat{Y} = \{y_1,\dots, y_S\}$ is defined as:
\begin{align}
p(\Rev|\vecbeta, \veclambda,\vec{\storeserror} , \errorterm) = \prod_{s=1}^{S} \mathcal{N}(y_s;\storerev,\errorterm),
\end{align}
where the model assumes constant-variance ($\errorterm$) for the Gaussian noise. \\

\paragraph{Prior Distributions:}
We assign prior distributions to all model parameters. First, we define a hierarchical prior distribution for $\vecbeta$, which we assume to be a Gaussian with mean $\vec{\mu}_\beta$  and covariance $\alpha^{-1} \imat$:
\begin{align*}
p(\vecbeta|\alpha) & = \mathcal{N}(\vecbeta; \vec{\mu}_\beta,\alpha^{-1} \imat),
\end{align*}
Following the standard  practices, we introduce a Gamma prior distribution with shape $\omega_1>0$ and scale $\omega_2>0$  for the hyper-parameter $\alpha$: 
\begin{align*}
p(\vect{\alpha}) & = \text{Gam}(\alpha;\omega_1,\omega_2)
\end{align*}
Similarly, we assign a Gamma prior  distribution with  shape $\rho_1$ and scale $\rho_2$ for the likelihood precision parameter $\gamma = \sigma^{-2}$: 
\begin{align*}
p(\gamma ) = \text{Gam}(\gamma;\rho_1,\rho_2),
\end{align*}
Finally, the following Gaussian prior distributions are selected for $\veclambda$ and $\vec{\storeserror}$ with mean  $\vec{\mu}$ and covariance $\vec{\varrho}\imat$, 
\begin{align*}
p(\veclambda) = \mathcal{N}(\veclambda;  \vec{\mu}_\lambda, \vec{\varrho}_{\lambda} \imat)\\
p(\vec{\storeserror }) = \mathcal{N}(\vec{\storeserror};  \vec{\mu}_\storeserror, \vec{\varrho}_{\storeserror} \imat).
\end{align*}\\[-0.7cm]

\paragraph{Posterior Distribution:}
The full vector of model parameters is denoted by $\vectheta = \{\vecbeta,\veclambda,\vec{\storeserror},\gamma\}$. Posterior probability given by:
\begin{align}
p(\vectheta|\dataset) = \frac{p(\dataset |\vectheta) p(\vectheta)}{ \int p(\dataset |\vectheta) p(\vectheta) \textrm{d} \vectheta }
\end{align}
where the marginal density takes the form:
\begin{align}
p(\dataset) =\idotsint p(\dataset|\vecbeta,\veclambda, \gamma)p(\vecbeta|\alpha)  p(\alpha) p(\veclambda)p(\vect{\storeserror}) p(\gamma)  \, \textrm{d} \vecbeta \, \textrm{d}{\alpha}  \, \textrm{d} \veclambda \, \textrm{d} \vect{\storeserror} \, \textrm{d} \gamma.
\label{equation: marginalData}
\end{align}

\begin{figure}[h!]
\centering
\tikz{ %
	\node[obs] (x_n) {$\indlocation_n$} ; %
	\node[latent, left =  of x_n]        		(sigma_s) {$\storesigma\imat$} ; %
	\node[obs, below = 2.4 cm of x_n]  			(x_s) {$\storelocation_s$}  ; %
	\node[latent, left = of  sigma_s]    		(store_error) {$\storeserror_s$}   ; %
	\node[obs, below = 2.4 cm of sigma_s]  		(store_features) {$\feat_s$} ; %
	\node[latent, below=  2.4 cm of store_error]        		(lambda) {$\veclambda$} ; %
	\node[tikprior,accepting, left = 1.5 cmof lambda]       (mu_lambda) {$.$} ; %
	\node[const, above=0.1 cm of mu_lambda]                 (mu_lambda_val) {$\mu_\lambda$} ; %
	\node[tikprior,accepting, left = 1.5 cm of store_error]	(mu_storeserror) {$.$} ; %
	\node[const, above=0.1 cm of mu_storeserror] (mu_storeserror_val) {$\mu_\storeserror$} ; %
	\node[tikprior,accepting, above =1cm of mu_lambda]     	(var_lambda) {$.$} ; %
	\node[const, above=0.1 cm of var_lambda] (var_lambda_val) {$\varrho_\lambda$} ; %
	\node[tikprior,accepting, above = 1cm of mu_storeserror]	(var_storeserror) {$.$} ; %
	\node[const, above=0.1 cm of var_storeserror] (var_storeserror_val) {$\varrho_\lambda$} ; %
	\node[obs, right=1cm of x_n]				        (phi_nq) {$\indfeat_{n}[-\indlocation_n]$} ; %
	\node[latent, right=of phi_nq]     					(beta) {$\vecbeta$} ; %
	\node[latent, right=of beta]        				(alpha) {\vect{$\alpha$}} ; %
	\node[tikprior,accepting, right= 1.cm of alpha]        		(alpha_a) {$.$} ; %
	\node[const, above=0.1 cm of alpha_a] (alpha_a_val) {$\omega_1$} ; %
	\node[tikprior,accepting, above= 1cm of alpha_a]        	(alpha_b) {$.$} ; %
	\node[const, above=0.1 cm of alpha_b] (alpha_b_val) {$\omega_2$} ; %
	\node[tikprior,accepting, above=1cm of alpha]        		(mu_beta) {$.$} ; %
	\node[const, above=0.1 cm of mu_beta] (mu_beta_val) {$\mu_\beta$} ; %
	\node[obs, below=2cm of phi_nq]  					(y_s) {$y_s$} ; %
	\node[latent, right= 3cm  of y_s] 					(gamma) {$\gamma$} ; %
	\node[tikprior,accepting, right=  1.cm of gamma] 			(gamma_a) {$.$} ; %
	\node[const, above=0.1 cm of gamma_a] (gamma_a_val) {$\rho_1$} ; %
	\node[tikprior,accepting, above = 1cm of gamma_a] 			(gamma_b) {$.$} ; %
	\node[const, above=0.1 cm of gamma_b] (gamma_b_val) {$\rho_2$} ; %
	\plate{plate1} {(beta)(phi_nq)} {P-2}; %
	\plate{plate4} {(lambda)(store_features)} {D-2}; %
	\plate{plate2} {(x_n)(phi_nq)(plate1.north west)(plate1.south west)} {N}; %
	\plate[inner sep=0.50cm, xshift=-0.12cm, yshift=0.12cm]{plate3} {(x_s)(x_n)(beta)(phi_nq)(y_s)(store_features)(lambda)(store_error)} {S}; %
	\edge {sigma_s} {x_n} ; %
	\edge {x_n,phi_nq} {y_s} ; %
	\edge {beta} {phi_nq} ; %
	\edge {alpha} {beta} ; %
	\edge {alpha_a} {alpha} ; %
	\edge {alpha_b} {alpha} ; %
	\edge {mu_beta} {beta} ; %
	\edge {gamma} {y_s} ; %
	\edge {gamma_a}{gamma}  ; %
	\edge {gamma_b}{gamma}  ; %
	\edge {lambda}{store_features}  ; %
	\edge {store_error}{sigma_s}  ; %
	\edge {mu_lambda}{lambda}  ; %
	\edge {var_lambda}{lambda}  ; %
	\edge {mu_storeserror}{store_error}  ; %
	\edge {var_storeserror}{store_error}  ; %
	\edge {x_s}{x_n}  ; %
	\edge {store_features}{sigma_s}  ; %
}
\caption{Plate diagram for the graphical representation for the \our.  Specifically, this express the spatial interaction between S number of stores with each store revenue $y_s$, located at $\storelocation_s$ with store features $\feat_s$ and N number of customers located at $\indlocation_n$ with P-2 characteristics $\indfeat_{n}[-\indlocation_n]$. We use Gaussian distributions as priors for $\vecbeta, \veclambda, \storeserror $ and Gamma distributions for $\gamma, \alpha$.  The diagram represents random variables with circles (\tikzcircle{5pt}), known values with grey filled circles (\tikzcirclefild{5pt}) while black filled circles (\tikzcircleblack{3pt}) indicate fixed parameters of prior and hyper-prior distributions, edges denote possible dependence, and plates denote replication.}
\label{fig:plate_diagram}
\end{figure}


\subsection{Inference}
Our goal is to estimate the posterior distribution over all parameters given the data  \ie $p(\vectheta|\dataset)$. Since marginal density is analytically intractable  (\eq~\ref{equation: marginalData}), we resort to approximate inference by employing two commonly used methods: Variational Inference (VI) \citep{Jordan1999} and  Markov Chain Monte Carlo (\mcmc) \citep{hastings1970monte}.\\[-1cm ]

\subsubsection{Variational Inference}
\vi is a powerful method to approximate intractable integrals where in contrast to \mcmc, it tends to be much faster because it rests on optimisation instead of sampling \citep{blei2017variational}. VI first posits a family of densities and then finds the member of that family, which is closest to the posterior by minimizing the Kullback-Leiber (\termkl) divergence.  Because the \termkl divergence cannot be directly calculated, alternatively, we maximise evidence lower bound, $\mathcal{L}_{\text{elbo}}$ that is equivalent to minimizing the \termkl divergence.\\

\paragraph{Variational Distributions:} We use the mean-field approximation and assumed a fully factorized variational distribution \citep{bishop2006pattern}: 
\begin{align}
q(\vecbeta,\alpha,\gamma,\veclambda,\storeserror) = q(\vecbeta) q(\alpha) q(\gamma)q(\veclambda)q(\storeserror),
\label{eq:full_posterior}
\end{align}
with 
\begin{align}
q(\vecbeta) & = \mathcal{N}(\vecbeta;\hat{\vec{\mu}_\beta},\vec{\Omega})
\label{eq:single_posteriors1}
\end{align}
\begin{align}
q(\alpha) & = \text{Gam}(\alpha;\hat{\omega_1} ,\hat{\omega_2}),\\
q(\gamma) & = \text{Gam}(\gamma;\hat{\rho}_1,\hat{\rho}_2),\\
q(\veclambda) & = \mathcal{N}(\veclambda; \hat{\vec{\mu}_\lambda},\vec{\mat{K}}_{\lambda}),\\
q(\vec{\storeserror}) & = \mathcal{N}(\vec{\storeserror}; \hat{\vec{\mu}_\storeserror},\vec{\mat{K}}_{\storeserror}),
\label{eq:single_posteriors2}
\end{align}

where $\vect{\nu} = \{\hat{\vec{\mu}_\beta}, \vec{\Omega}, \hat{\omega_1},\hat{\omega_2},\hat{\rho}_1,\hat{\rho}_2,\hat{\vec{\mu}_\lambda},\vec{\mat{K}}_{\lambda},\hat{\vec{\mu}_\storeserror},\vec{\mat{K}}_{\storeserror}\}$ are the variational parameters which are optimized within the algorithm. \eqs~\eqref{eq:single_posteriors1}--\eqref{eq:single_posteriors2} define our approximate posterior. With this, we give details of the variational objective function, \ie $\elbo$, which we aim to maximize with respect to $\vect{\nu}$.\\

\paragraph{Evidence Lower Bound:} Following the standard variational inference, \elbo can be written as a combination of expected log likelihood $(\mathcal{L}_{\text{ell}})$ and \termkl-divergence term $(\mathcal{L}_{\text{kl}})$:
\begin{align}
\mathcal{L}_{\text{elbo}}(\vect{\nu}) = \mathcal{L}_{\text{ell}}(\vect{\nu}) - \mathcal{L}_{\text{kl}}(\vect{\nu}).
\label{eq:elbo}
\end{align}
the expected log likelihood term can be written as
\begin{align}
\mathcal{L}_{\text{ell}} & = \mathbb{E}_{\vecbeta,\gamma,\veclambda, \storeserror} [\ln p(Y|\vecbeta, \gamma,\veclambda, \vec{\storeserror})] \nonumber \\[5pt]
& =  -\frac{S}{2}\ln 2\pi + \frac{S}{2}(\psi(\hat{\rho}_1)-\ln{\hat{\rho}_2}) - \\ \nonumber
& \qquad \frac{1}{2} \frac{\hat{\rho}_1}{\hat{\rho}_2}\mathbb{E}_{\vecbeta,\gamma,\veclambda,\vec{\storeserror}}\left[\frac{\gamma}{2} \sum_{s=1}^S \left( y_s -  \vecbeta^\top\sum_{n=1}^N\indfeat_{n}[-\indlocation_n] \frac{\mat{Z}_s(\indlocation_n)}{\sum_{j=1}^{S} \mat{Z}_j(\indlocation_n)}\right)^2\right]
\end{align}
the \termkl-Divergence Term is expanded and simplified as: 
\begin{align}
\mathcal{L}_{\text{kl}} & = \mathbb{E} [\ln p(\vectheta)] - \mathbb{E} [\ln q(\vectheta)] \nonumber\\
& =  \mathbb{E}_{\vecbeta,\alpha} [\ln (p(\vecbeta | \alpha)] + \mathbb{E}_{\alpha} [\ln p(\alpha)]+  \mathbb{E}_{\veclambda} [\ln p(\veclambda)] +\mathbb{E}_{\storeserror} [\ln p(\storeserror)] + \mathbb{E}_{\gamma} [\ln p(\gamma)] - \\
& \qquad  \mathbb{E}_{\vecbeta} [\ln q(\vecbeta)]- \mathbb{E}_{\alpha} [\ln q(\alpha)]-\mathbb{E}_{\veclambda} [\ln q(\veclambda)] -\mathbb{E}_{\storeserror} [\ln q(\storeserror)] - \mathbb{E}_{ \gamma} [\ln q(\gamma)] \nonumber,
\end{align}
where each term is given in the Appendix A. $\mathcal{L}_{\text{elbo}}(\vect{\nu})$ is not computable in analytically closed forms and remains intractable. Hence we resort to  Black Box variational inference method where the gradient is computed from the Monte Carlo samples from the variational distributions \citep{ranganath2014black}.  We implement the algorithm using Tensorflow 2 \citep{abadi2016tensorflow} in Python 3. \\[-1cm]

\subsubsection{Markov Chain Monte Carlo}
In order to compare our estimations we describe the \mcmc which has been the dominant paradigm for approximate inference for decades. First, we construct a Markov chain on $\vectheta$  whose stationary distribution is the posterior $p(\vectheta|\dataset)$. Then we collect samples from the stationary distribution by sampling from the Markov chain. Finally, we use the collected samples to approximate the posterior with an empirical estimate. \mcmc methods ensure producing exact samples from the target density but tend to be computationally intensive  \citep{robert2013monte}. When the datasets are large, \mcmc becomes slower and computationally expensive to form inferences.  We use open-source software, Stan which is a C++ library for Bayesian modeling, with the R interface to compile results \citep{rstan}. We adopt the No-U-Turn sampling method, an extension to Hamiltonian Monte Carlo algorithm for the experiments \citep{hoffman2014no}.

\subsection{Edge Correction}
\begin{figure}[h]
\centering
\includegraphics[width=6cm]{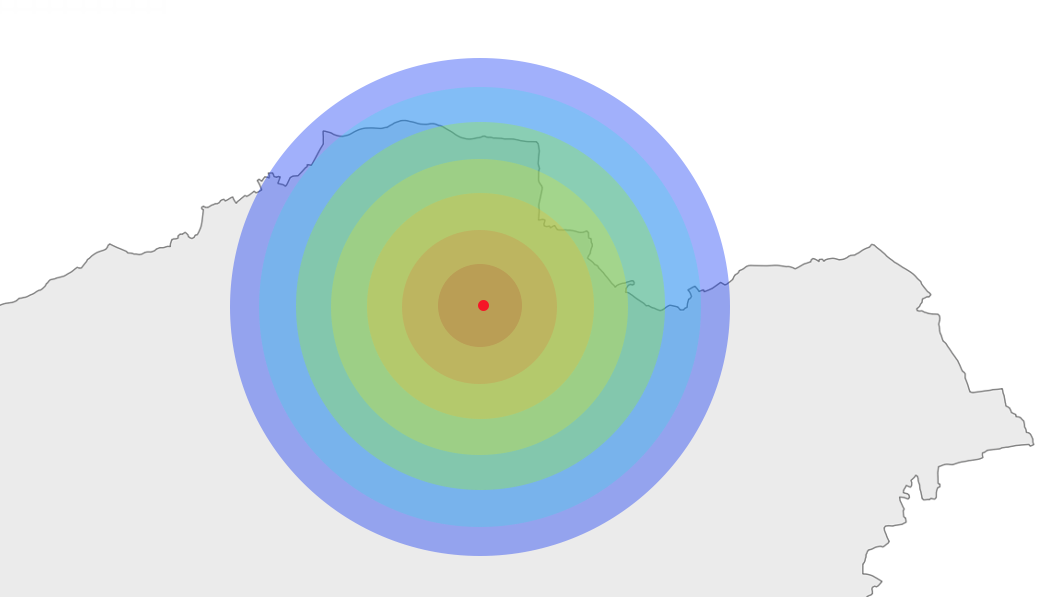}\\		
\caption{ The red marker denotes a store at the edge of London. There may be customers who contributes to its revenue but not in the study area. Intersection of the radius and London map results in an arbitrary polygon shape.}
\label{fig:edge_correction}
\end{figure}

Stores on the edge of the study area $\region$ cannot be evaluated without a certain bias because the model cannot capture the contribution from customers living outside $\region$. To overcome this, we adjust the revenues of the stores $\{y_s\}_{s=1}^S$,  and this is carried out before fitting the model. Following a similar approach to the model, we assume a Gaussian centered on the store and calculate the area under the curve (\auc)  $\mathcal{A}$, which intersects with the study area. We set the variance $\eta^2$ of the Gaussian to be $\maxdist/4$ to cover approximately an area of 0.99 within the buffer radius of $\maxdist$ around the store center $\storelocation_s$. Calculating the \auc for an arbitrary polygon as shown in \fig~\ref{fig:edge_correction},  is computationally challenging. Henceforth we use the Monte Carlo method, where the samples are drawn from $\mathcal{N}(\storelocation_s, \eta^2 \imat)$ and reject them if outside the $\region$  to calculate the fraction of kept samples. 

We formulate the adjusted revenue $\widetilde{y_s}$ as the actual revenue weighted by the \auc:
\begin{align}
\widetilde{y_s}  = y_s \times  \mathcal{A}.
\label{eq:edge}
\end{align}
We apply this to real-world data for edge correction before fitting the \our.

\section{Simulation study}
\label{section:simulations}
\begin{figure}[h!]
\centering
	\begin{minipage}{.4\linewidth}
	\centering
	\includegraphics[width=5cm]{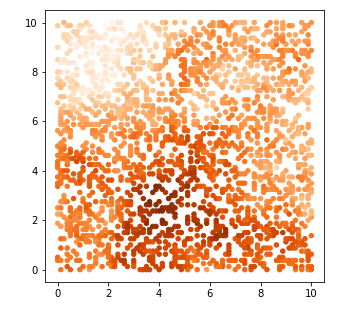}\\		
	\scriptsize (a) 
	\end{minipage}%
	\begin{minipage}{.4\linewidth}
	\centering
	\includegraphics[width=5cm]{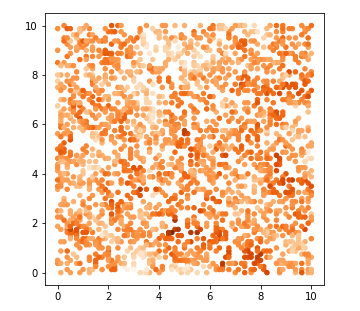}\\
	\scriptsize	(b) 
	\end{minipage}
\caption{Simulated Customer features for $N = 1000$ under two different spatial correlation structures to closely simulate the real-world scenarios: (a) Strong Spatial Correlation; (b) Moderate Spatial Correlation. }
\label{fig:cust_features}
\end{figure}

We design a simulation study to examine the inferences obtained from \vi and \mcmc methods under different synthetic settings characterised by an increasing number of stores and customers. We also compare the computational performance of the two methods by observing the run time of each fitted model. First, we simulate the data from a spatial process that closely matches the modeling framework introduced in Section \ref{section:methodology}, \eq~\eqref{eq:model2} with $\storeserror_s = 0 $. The process is defined as:
\begin{align}
y_s|\vect{\beta}, \vect{\lambda}, \sigma^2 \sim \mathcal{N}(\storerev, \errorterm),
\end{align}
where the locations of stores and customers are simulated within a square. Two customer features are generated, one with a strong spatial correlation and the other with a moderate spatial correlation to closely reflect the real-world customer features as shown in  \fig~\ref{fig:cust_features}. The store locations are randomly sampled within the same spatial boundaries used to sample the customers. Store features are sampled from a Gamma distribution ($\capitalfeat \sim \text{Gam}(1,1) $) to represent features such as floorspace. 
 
\vspace{-0.5cm}
\subsection{Parameter Estimation}
For both \vi and \mcmc methods, all priors are chosen to be weakly informative to allow the data to drive the inference as illustrated in Table~\ref{tab:post_dist}. We fit our \mcmc model using one chain with 5000 iterations by removing the first 2500  for warm-up, and every post-warm-up iteration is used for posterior samples. The posterior distributions along with the prior distributions are visualised in  Table~\ref{tab:post_dist} and parameter estimates are presented in Table~\ref{tab:parameter_estimates}.  The results indicate that both methods approximate the posterior mean effectively and variational approximations of the posterior variance are lower than \mcmc method.  

\begin{table}
\caption{The first row indicates the True values of the parameters used to create the synthetic data, and the following rows display the first (Mean) and second moments (Standard deviation) along with its 95\% quantile-based Credible Intervals (CI) for the posterior distributions for \vi and \mcmc methods. \label{tab:parameter_estimates}}
\fontsize{10}{12}\rm\tabcolsep=0.03cm
\centering
\begin{tabular}{ccccccc}
\toprule
	&      & $\beta_1$  & $\beta_2$   & $\lambda_1$    & $\lambda_2$ & $\gamma$  \\  
	\midrule
	True                           &          & $-0.2$   & 0.4 & 0.1  & 0.5 & 4       \\
	\multirow{3}{*}{\vi}      & Mean &  $-0.196$  &  0.398  &   0.164 &  0.383  &   1.821   \\
	& Std    &  0.014   & 0.018   &  0.235   & 0.116   &  0.727    \\
	& CI      &   ($-0.224, -0.169$) &  (0.362,0.434)    &  ($-0.296,0.625$)  &   (0.156, 0.609) &   (0.687, 3.499)   \\
	\multicolumn{1}{c}{\multirow{3}{*}{\mcmc}} & Mean &  $-0.198$    & 0.400    &  0.185  &  0.387  &    1.908  \\
	\multicolumn{1}{c}{}                      & Std  &   0.017 &   0.021  &   0.547 &  0.393   &  0.904    \\
	\multicolumn{1}{c}{}                      & CI   &  ($-0.235, -0.166$)  & (0.358 , 0.447)   &   ($-0.839$, 1.387 )  &   ($-0.313$,  1.269) &  (0.562, 4.054)   \\ \bottomrule
\end{tabular}
\end{table}

\begin{table}
	\caption{Column one demonstrates the weakly informative prior distributions, and the following columns illustrate marginal posteriors of the interested parameters inferred by \vi and \mcmc. Synthetic experiment consists of 10 stores and 1000 customers ($S = 10$, $N = 1000$). \label{tab:post_dist}}
\centering
\rm\tabcolsep=0.07cm
\begin{tabular}{ccc}  
	\toprule
	Prior & \multicolumn{2}{c}{\vi vs. \mcmc}\\
	\midrule  
 	$p(\vect{\beta}|\alpha) \sim  \mathcal{N}(0,\alpha^{-1}\imat )$ & $\beta_1$  & $\beta_2 $ \\
 	$p(\alpha) \sim  \Gamma(1,1) $ & & \\
 	\includegraphics[scale=0.3]{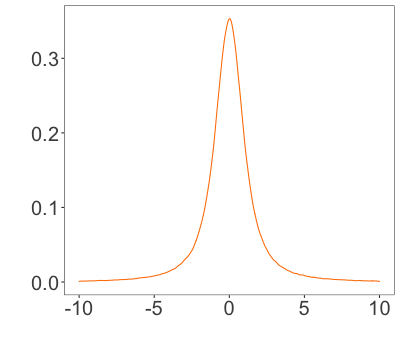}& 
	\includegraphics[scale=0.3]{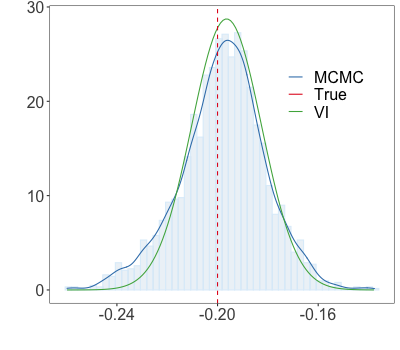}&
	\includegraphics[scale=0.3]{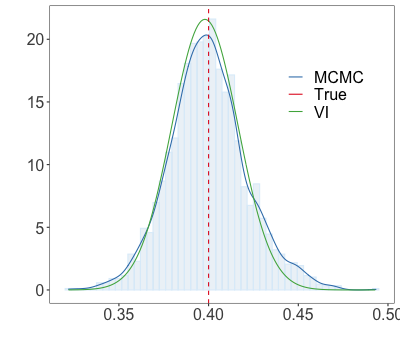}\\ 
	$p(\lambda) \sim  \mathcal{N}(0,\alpha^{-1}\imat)  $ &$\lambda_1$ &  $\lambda_2$\\
	\includegraphics[scale=0.3]{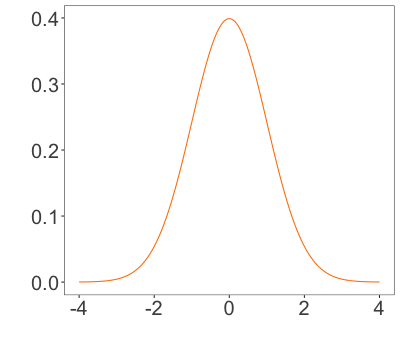}&
	\includegraphics[scale=0.3]{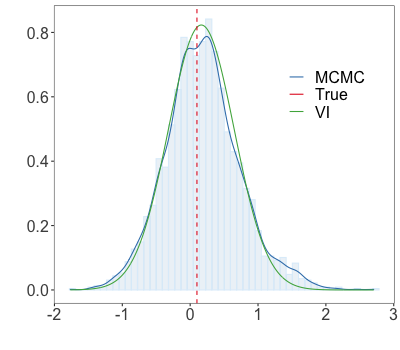}&
	\includegraphics[scale=0.3]{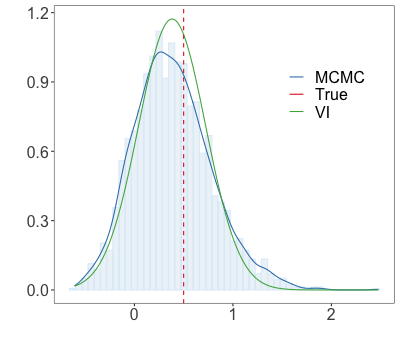}\\	
		$p(\gamma) \sim  \Gamma(1,1)  $ & $\gamma $ & \\
	\includegraphics[scale=0.3]{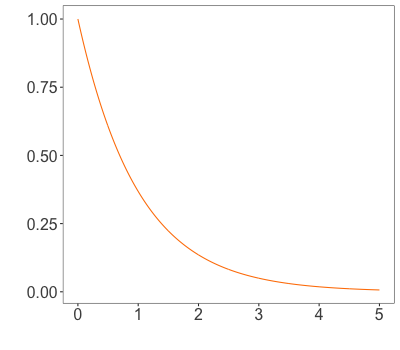}&
	\includegraphics[scale=0.3]{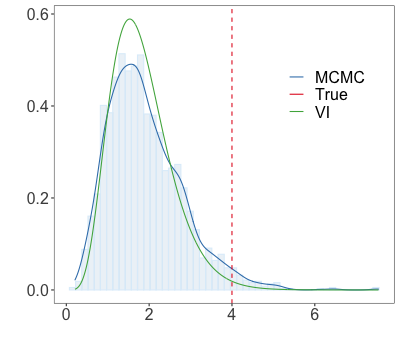}&\\	
	\bottomrule
	\end{tabular}
\end{table}

The simulation process explained above is experimented under two different synthetic settings:
\begin{enumerate}
	\item $sim_1$: 10 stores with 1000 individuals ($S = 10, N = 1000$)
	\item $sim_2$: 50 stores with 2000 individuals ($S = 50, N = 2000$)
\end{enumerate}

We simulate random store locations to create 50 datasets and compare the performance across datasets using the posterior means of $\bm{\beta},\lambda, \gamma $ and the 95\% quantile-based credible intervals for each parameter from each fitted model. Three standard measures are used to compare the performance between \mcmc  and \vi methods: 
\begin{enumerate}
	\item  the bias, which measures the differences between the posterior mean from the model fit to $\text{dataset}_i$ ($\hat\beta_{i} $) and the true value of the parameter $\beta$, Bias = $\frac{1}{50}  \sum_{i=1}^{50}(\hat\beta_{i} - \beta)$;
	\item the mean-squared error (\mse), which takes the squared of the difference between posterior mean and true value, \mse = $\frac{1}{50}  \sum_{i=1}^{50}(\hat\beta_{i} - \beta)^2$;
	\item the coverage of the 95\% quantile-based credible interval obtained from fitting the model to $\text{dataset}_i$, coverage = $\frac{1}{50} \sum_{i=1}^{50} I \left (\beta \in \text{credible interval}_i \right) $, where $I(\cdot)$ is the indicator function equal to 1 if the statement is true and 0 otherwise.
\end{enumerate}

Table~\ref{tab:performance_synth1} and Table~\ref{tab:performance_synth2} show the results of the fitted models for the two synthetic settings, averaged across the 50 datasets. Both \vi and \mcmc algorithms exhibit comparable performance in terms of bias, \mse, and coverage across both simulation studies. For $sim_1$, we observe lower coverage for $\gamma$ with the \vi scheme. However the coverage for $\gamma$ is improved to one in the $sim_2$.  Both $\lambda$ and $\gamma$ parameters result in a higher estimated \mse under both the simulation setting for \vi and \mcmc methods. This is an indication of the lack of identifiability in the parameters due to the flexibility in the model. The precision $\gamma$  of the error term $\errorterm$ tends to be underestimated on average. Both models are fitted on a Intel Xeon CPU (3.5GHz and 32 GB of RAM). The run time of the \vi algorithm is about five times faster than the \mcmc algorithm in the simulation study. This is vital for our real-world data application, where the number of spatial locations is much larger than the synthetic settings. Overall the \vi algorithm exhibited a reduced run time while providing good estimations and inference of the parameters of interest in this simulation study. 

\begin{table}
\caption{ \vi and \mcmc simulation study performance for $S = 10, N = 1000$. \label{tab:performance_synth1}}
\rm\tabcolsep=0.4cm
\begin{tabular}{llrrrrr} 
	\toprule
	Metric & Method &  $\beta_1$  & $\beta_2$  & $\lambda_1$  & $\lambda_2$  & $\gamma$ \\
	\midrule
	\multirow{2}{4em}{Bias}		 	& \vi & $-0.00$2 &	0.004 &	0.258 & 0.110 &	$-1.828$\\ 
						    		& \mcmc & $-0.002$ &0.004 &	 0.265  &	0.116 & $-1.772$ \\ 
	\multirow{2}{4em}{\mse} 			& \vi & 0.000 & 0.000 &	0.130 &	0.051 &	3.467\\ 
							   		& \mcmc & 0.000   &	0.42   &	0.130  &0.049 &3.276 \\ 
	\multirow{2}{4em}{Coverage} 	& \vi & 0.94  & 0.96 &	1. &	1. &	0.44 \\ 
						        	& \mcmc & 0.96 &	0.98 &	1. &	1. &	0.94 \\ 
   	\midrule
	\multicolumn{2}{c}{\multirow{2}{*}{Run time (s) }}& \multicolumn{2}{c}{\vi}  &  \multicolumn{2}{c}{\mcmc} \\
	\multicolumn{2}{c}{} 	      &\multicolumn{2}{c}{207} &  \multicolumn{2}{c}{1064} \\
	\bottomrule
\end{tabular}
\end{table}

\begin{table}
\caption{\vi and \mcmc simulation study performance for $S = 50, N = 2000$.	\label{tab:performance_synth2} }
\rm\tabcolsep=0.4cm
\begin{tabular}{llrrrrr} 
	\toprule
	Metric & Method &  $\beta_1$  & $\beta_2$  & $\lambda_1$  & $\lambda_2$  & $\gamma$ \\  \midrule
	\multirow{2}{4em}{Bias} & \vi & 0.000  & 0.002  &$-0.338$ & 0.352   & $-0.754$	\\ 
 						  	& \mcmc &  0.002&-0.001   & $-0.092$&	 0.341 &$-0.734$ \\ 
	\multirow{2}{4em}{\mse} 	& \vi & 0.000 &  0.000  &	0.185     &	0.179   &  0.598 \\ 
      						& \mcmc & 0.000 & 0.000 & 0.008 & 0.186 & 0.571 	    \\ 
	\multirow{2}{4em}{Coverage} & \vi & 1. & 0.94 &	0.84 & 0.94 & 1. \\ 
						   	& \mcmc & 1. & 1. & 1. & 0.857 & 1. \\ 
   	\midrule
	\multicolumn{2}{c}{\multirow{2}{*}{Run time (s) }}& \multicolumn{2}{c}{\vi}  &  \multicolumn{2}{c}{\mcmc} \\
	\multicolumn{2}{c}{} 	      &\multicolumn{2}{c}{1079} &  \multicolumn{2}{c}{5280} \\
	\bottomrule
\end{tabular}
\end{table}


\subsection{Model Comparison}

Finally under the simulation study, we conduct a comparison of our model with the Huff modified model \citep{li2012assessing}. Two standard metrics are used to evaluate the performance: 
\begin{enumerate}
	\item the Normalised Root-Mean-Squared Error (\NRMSE), which measures the differences between the values predicted by a model ($\predRev$) and the values observed ($\Rev$), $\NRMSE=\frac{\sqrt{E[\Rev - \predRev ]^2 }}{E[\Rev]}$; 
	\item the R-squared, which is the ratio of the variance of the residuals ($\text{SS}_{res}$) and he variance of the observed $\Rev$ ($\text{SS}_{tot}$), $\Rsqd = 1 - \frac{\text{SS}_{res}}{\text{SS}_{tot}}$.
\end{enumerate}	

Table~\ref{tab:performance_huf} displays the results for \our and Huff modified model. \our exhibits better performance across both the settings compared to the modified Huff model. We observe an increase in \NRMSE for both models as the number of stores and customers increases. However, the $\Rsqd$ remains unaffected at significantly high levels showing more robust performance for \our under both simulation settings compared to the modified Huff model.  

\begin{table}
\caption{Performance of the simulation studies for \our and Huff modified model. $sim_1: S = 10, N=1000$ and $sim_2: S = 50, N=2000$ \label{tab:performance_huf}.}
\rm\tabcolsep=0.5cm
\begin{tabular}{llcc}
	\toprule
	&              & $sim_1$            & $sim_2$       \\[5pt]
	\midrule
	\multirow{2}{*}{$\Rsqd$}   	& Model         &     0.98      &      0.94          \\
							   	& Modified Huff Model &     0.77      &    0.30            \\
	\multirow{2}{*}{NRMSE}  	& Model         &    0.07       &    0.15        \\
								& Modified Huff Model &      0.24   &        0.64         \\
	\bottomrule
\end{tabular}
\end{table}

\section{Large-scale geospatial dataset of pub activities in London}
\label{section:data}
We initially develop a large-scale, geospatial dataset for England non-domestic properties using data from multiple sources. However, for the interest of this study, we limit to one non-domestic property category, public houses (pubs), which are located in Greater London. To the best of our knowledge, this is the first study exploring these datasets together to benefit retail businesses. A detailed description of each dataset is given in  Appendix B.  All the spatial data processing is done using PostGIS on a PostgreSQL database.

\subsection{Store level data}
\begin{figure}[h!]
\centering
	\includegraphics[width=14cm]{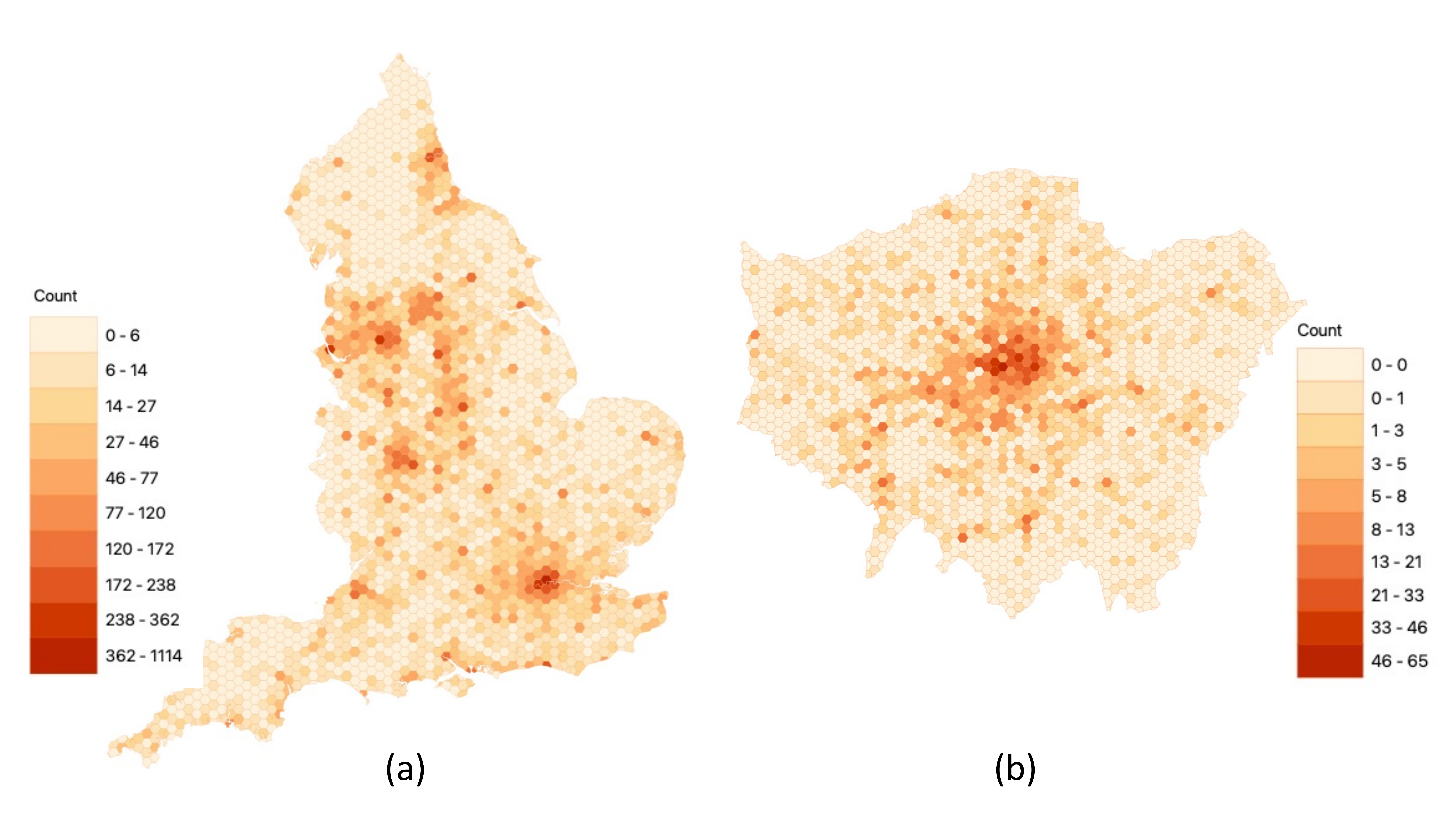}
\caption{Spatial distribution of pubs: (a) across England; (b) zoomed into Greater London. The region is split into equal size grids of hexagons (size of each side : (a) 5km; (b) 0.5km) and number of pubs within each hexagon is displayed with a colour gradient. }
\label{fig:pubsEng}
\end{figure}

We compile a dataset with stores' geospatial location, rateable values, and store-specific features. We do this by Joining the VOA \citep{voa} and Addressbase from \citet{osAb} data using the cross-reference which renders all of the non-domestic properties geo-coordinates and their rateable values. The calculation of the rateable values of  pubs is different from other categories. In contrast, the rateable value of pubs is based on the annual level of trade (excluding VAT) that a pub is expected to gain if operated in a reasonably efficient way \citep{voaPubs}. Hence the rateable value is a good proxy of the pub revenues, and we use data related to pubs for the real-world experiment in this study. There are 40,000  pubs recorded in VOA for England and Wales. The spatial distribution of pubs across England is shown in \fig~\ref{fig:pubsEng}. 

\begin{figure}[h]
\centering
\fbox{\includegraphics[width=13.5cm]{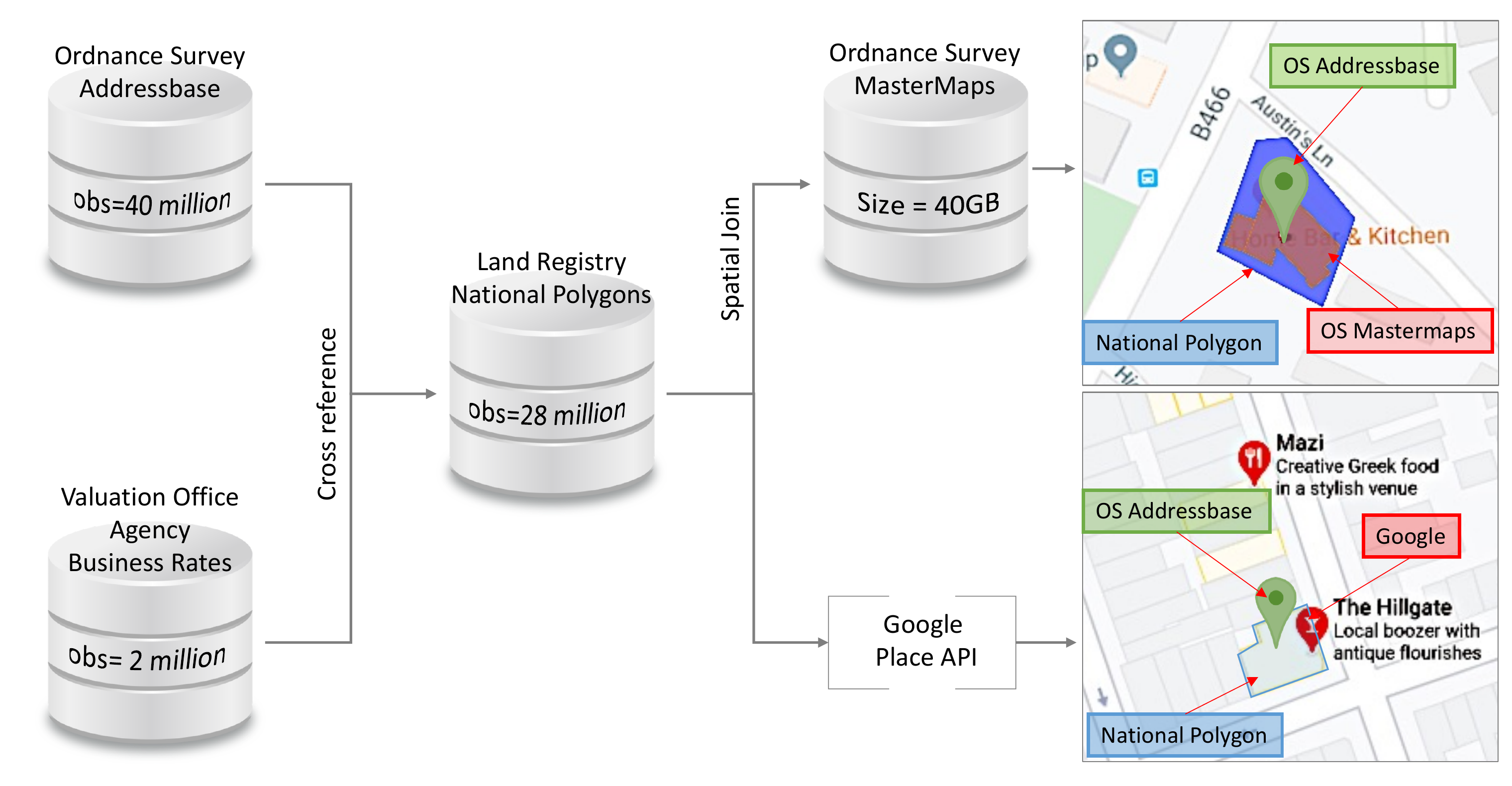}}
\caption{Diagram illustrates the steps to extract the store features. Each dataset is named as per the data source along with its number of records (obs) or size. Initially OS addressbase is joined with VOA dataset and then spatially joined with National Polygons data to find the Title polygon of each land. This is next joined with Mastermaps and linked with Google data to obtain the store footprints and google customer ratings respectively.}
\label{fig:flowdiagram}
\end{figure}

The store features are an essential factor in assessing the attractiveness of the stores. The internal store characteristics of the building, such as the floor size, height are extracted from the OS Mastermaps from \citet{osMmData}. This is accomplished by first spatially joining the polygon of the land \citep{nationalPolygons} with locations of stores and next spatially join the polygon of the footprint from Mastermaps.  Additionally, external characteristics such as the closest distance to public transport access points \citep{NaPTAN}, tourist attractions \citep{historicengland} are calculated using the Euclidean distance between spatial locations. We have strengthened the store attractiveness measures by using the customer reviews on Google \citep{Google2020}.  People can write reviews and rate the places voluntarily on Google maps. The ratings are then aggregated and shown to the public. Using the Google Places API, this data can be accessed at a cost. Flow diagram of the process used to extract the store features are demonstrated in \fig~\ref{fig:flowdiagram}.

\vspace{-1cm}
\subsubsection{Customer level data}
The most granular level of customer data can be identified as the residential locations. OS Addressbase dataset provides both residential and commercial addresses (over 40 million) along with geo-locations. However, since there is no data for customer features at the residential level,  in this study, we use postcodes which is the next most granular level. Henceforth, we assume that the customers' behaviour who are residing in the same postcode are homogeneous. In Greater London on average there are 17 households per postcode. The postcode centroids for Greater London are displayed in \fig~\ref{fig:pubslondon}. The population and proportion of gender at the postcode level are used to reflect the demographics in the area.
Additionally, we employ the deprivation data to understand the customer characteristics in the area \citep{deprivationData}. There are seven domains of deprivation categories: (1) Income Deprivation, (2) Employment Deprivation, (3) Education, Skills and Training Deprivation, (4) Health Deprivation and Disability, (5) Crime, (6) Barriers to Housing and Services and (7) Living Environment Deprivation. Deprivation level data is provided at the LSOA level. We assign that to the postcodes by point to polygon spatial join.\\

\vspace{-0.7cm}	
\section{Case study: estimating revenues of pubs in London}
\vspace{-0.1cm}	
\label{section:casestudy}
In this section,  we illustrate our proposed methodology using the pubs' dataset developed for Greater London in section \ref{section:data}. After compiling data from different sources, the final complete dataset consists of $S = 1804$ pubs. The derived approximated revenue after adjusting for edge correction (\eq~\eqref{eq:edge}) is used as the response variable $y_s$ in the model with natural log transformation. For each pub, we derived pub-specific features: floorspace, height, number of floors, the total area of land; distance to the closest metro, train station, bus stop, park, popular attractions, sports facility; customer rating on Google, number of users rated and an indicator to show if the pub is in a major town.

\begin{figure}[h!]
\centering
	\begin{minipage}{.5\linewidth}
	\centering
	\includegraphics[width=6.5cm]{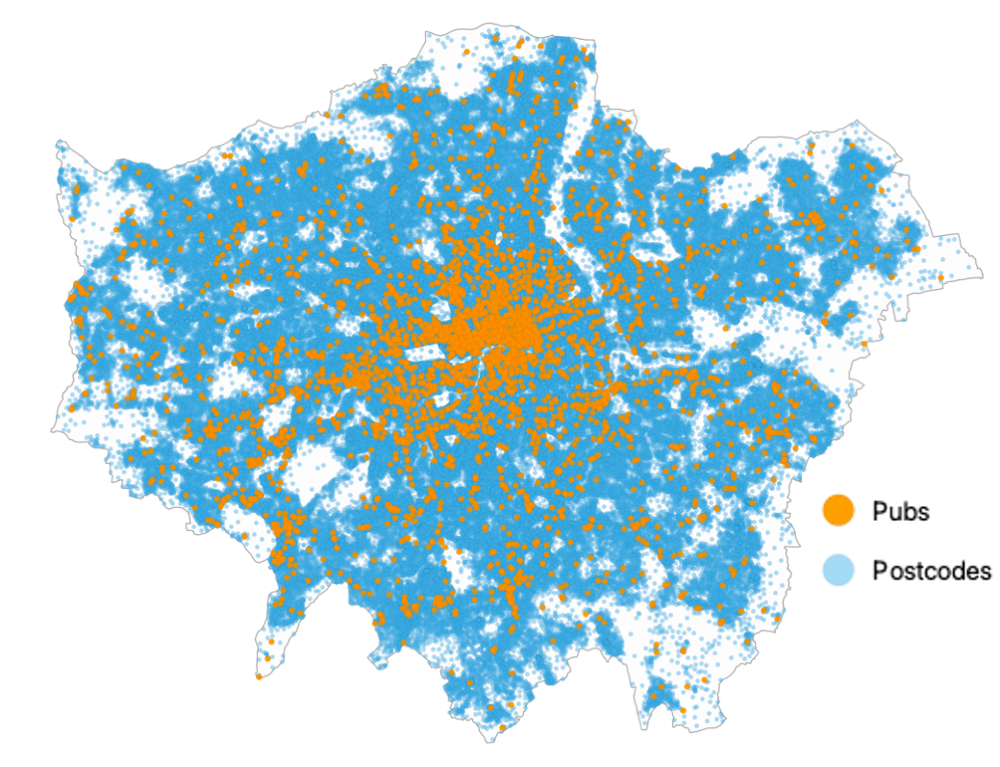}\\
	\scriptsize (a) 
	\end{minipage}%
	\begin{minipage}{.5\linewidth}
	\centering
	\includegraphics[width=6.5cm]{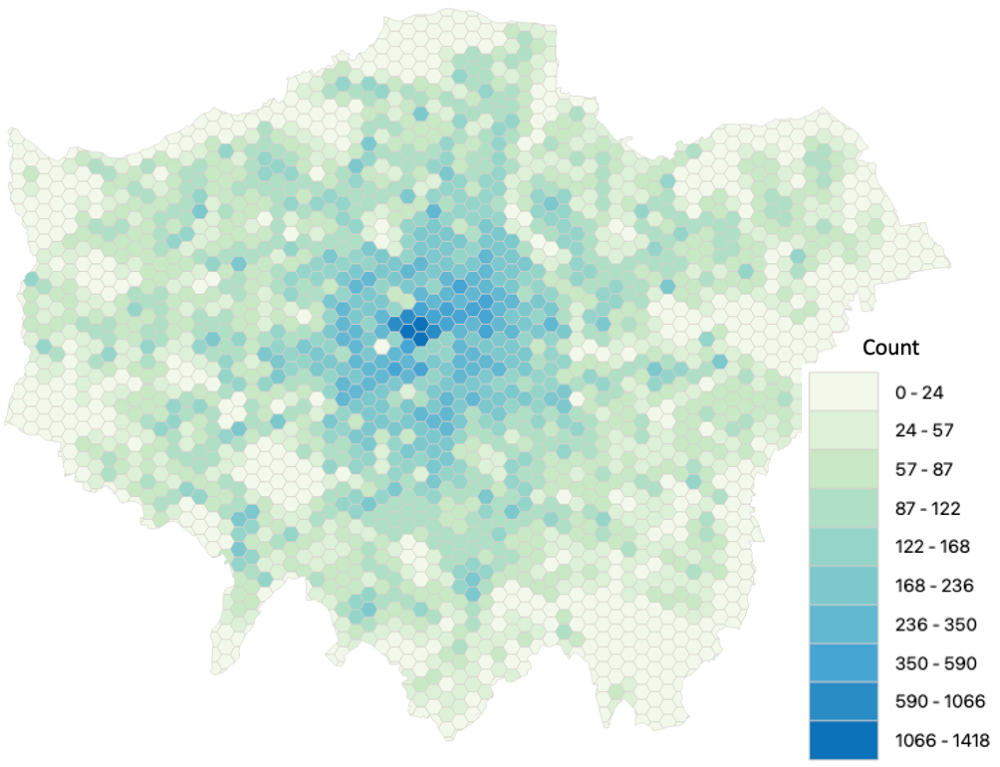}\\ 
	\scriptsize (b) 
	\end{minipage}
\caption{(a) Visualization of the locations of pubs in orange markers ($S = 1804$) and postcode centroids in blue markers ($N = 174360$) over the map of London; (b) Greater London is split into equal size grids of hexagons (size of each side is 0.5km) and number of postcodes within each hexagon is displayed with a colour gradient. }
\label{fig:pubslondon}
\end{figure}

We determine the customer locations at the postcode level, which is the most granular level of census estimates are released. There are $N = 174360$ postcodes for Greater London. We represent the characteristics of the postcodes by the population at each postcode and its proportion of male, and  deprivation scores. All features have been normalized before training the model. The model may be improved with more granular customer-specific characteristics; underlying arguments would remain the same. Centroids of the postcodes and retail locations of the pubs are presented in \fig~\ref{fig:pubslondon}(a), on a map of London.\\[-0.5cm] 
 
\begin{figure}
\centering
	\begin{minipage}{.33\linewidth}
 	\centering
 	\includegraphics[width=4cm]{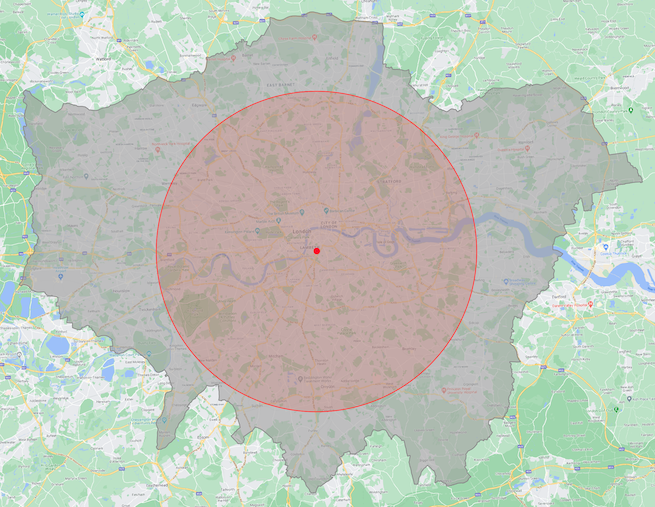}\\
 	\scriptsize (a)
 	\end{minipage}%
 	\begin{minipage}{.33\linewidth}
 	\centering
 	\includegraphics[width=4cm]{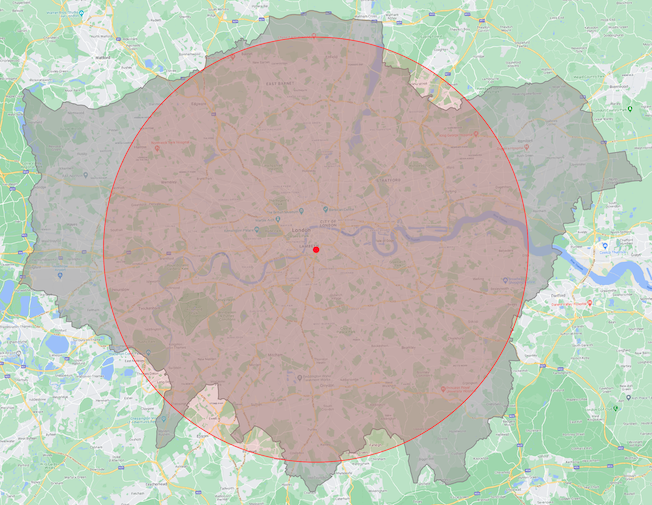}\\
 	\scriptsize (b)
 	\end{minipage}%
 	\begin{minipage}{.33\linewidth}
 	\centering
 	\includegraphics[width=4cm]{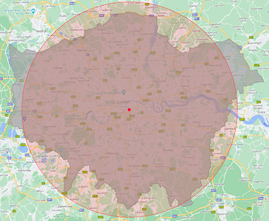}\\
 	\scriptsize (c) 
 	\end{minipage}
\caption{Demonstration of different radius used for truncated Gaussian with an example concerning a pub located in the center of London. Three radii were used in the study: (a) 15km; (b) 20km; (c) 25km.}
\label{fig:TruncRadiusMap}
\end{figure}

Customer behavior is not affected after a certain distance from the business facility, despite the pubs' attractiveness. We explore the model under three different radius, $\maxdist =$  15km, 20km and 25km  as presented in \fig~\ref{fig:TruncRadiusMap}.  We calculate the distance between origin and pub using Euclidean distance, although a better representation would use a transport network. 
 
 We first perform a preliminary study of our model with a store-specific coefficient which denotes the store-specific variance $\sigma^2_s = \exp ( \upsilon_s )$, representing the attractiveness of the store as given by \eq~\eqref{eq:model1}. We experiment with the model for three different radii of the truncated Gaussian and model performance summarised in Table~\ref{tab:model_perfomance}. Results indicate that $\Rsqd$ increased to 0.72 as the radius increased from 15km to 20km but reduced to 0.57 as the radius increased to 25km. Hence the best experimental results yielded for truncated Gaussian with a radius of 20km.
 
\addtolength{\tabcolsep}{15pt} 
\begin{table}
\caption{ $\Rsqd, \errorterm$  and  \NRMSE for the fitted \our with revenues of pubs in Greater London under three different radii of the truncated Gaussian. \label{tab:model_perfomance} }
\rm\tabcolsep=0.7cm
\begin{tabular}{lccc}
	\toprule
    				& \multicolumn{3}{c}{Truncated radius (km)} \\ 
    \midrule
    				& 15  	     &     20        &      25      \\
    \midrule
	$\Rsqd$           &       0.19  	   &   0.72 	  &     0.57   \\
	$\errorterm$ 	&       0.67   	  &    0.45       &      0.52  \\
    \NRMSE   		&       0.08    	  &     0.05      &      0.06      \\
    \bottomrule
\end{tabular}
\end{table}
\addtolength{\tabcolsep}{-15pt}

\begin{figure}[h!]
\centering
	\begin{minipage}{.5\linewidth}
	\centering
	\includegraphics[width=7cm]{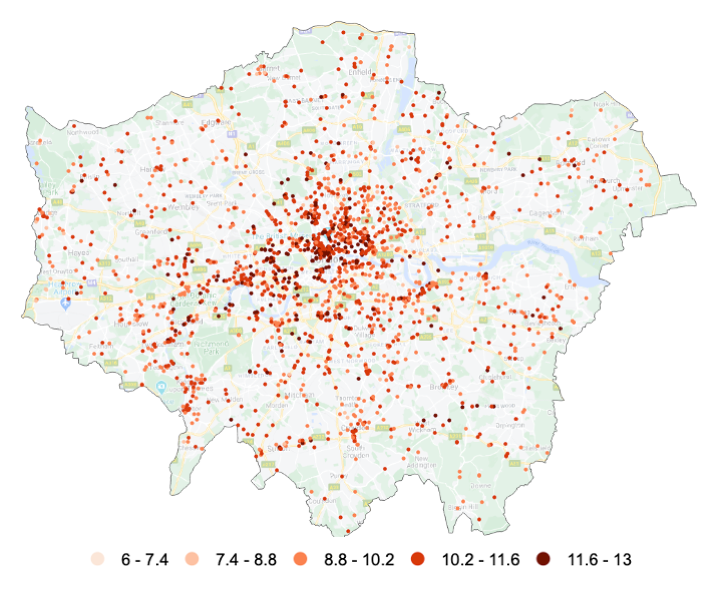}\\
	\scriptsize (a) 
	\end{minipage}%
	\begin{minipage}{.5\linewidth}
	\centering
	\includegraphics[width=7cm]{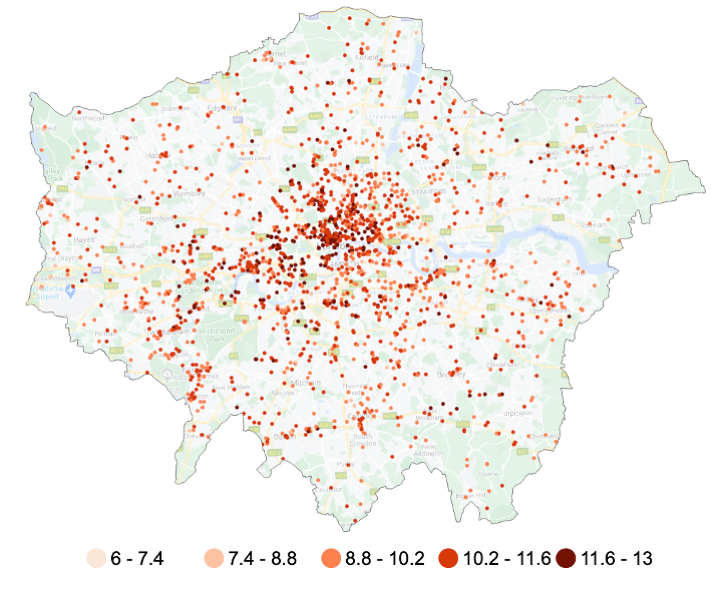}\\ 
	\scriptsize (b) 
	\end{minipage}
	\begin{minipage}{.5\linewidth}
	\centering
	\includegraphics[width=7 cm]{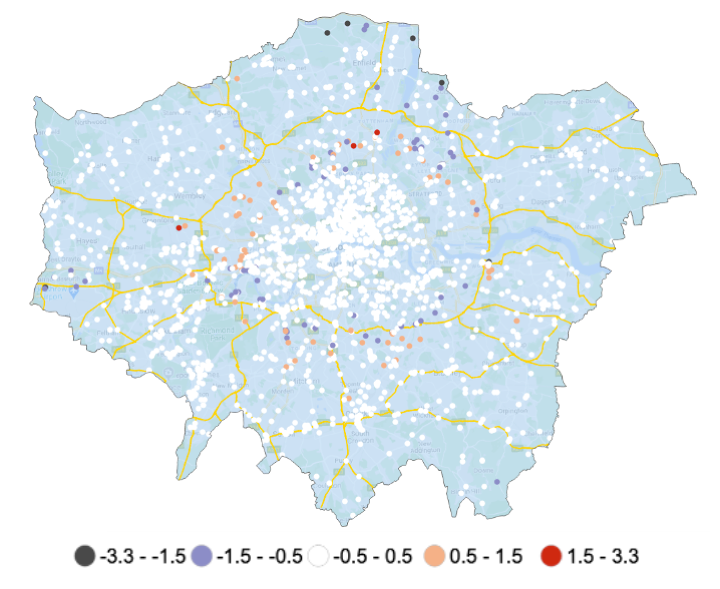}\\ 
	\scriptsize (c) 
	\end{minipage}%
	\begin{minipage}{.5\linewidth}
	\centering
	\includegraphics[width=7cm]{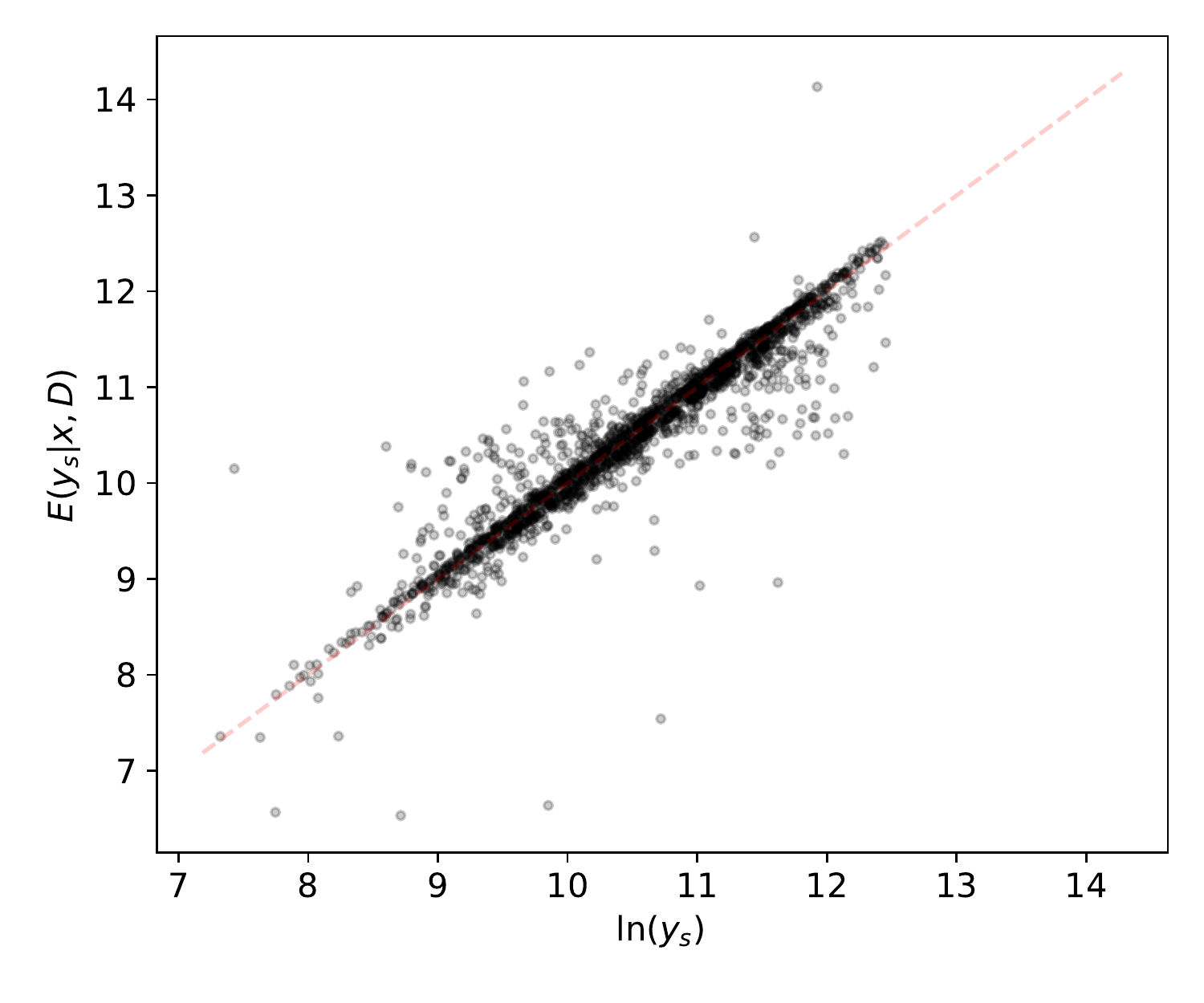}\\ 
	\scriptsize (d) 
	\end{minipage}
\caption{Visualisation of the Pub's revenue and predictions over greater London map with truncated Gaussian radius of 20 km:  (a) Revenue at each pub; (b) Predicted revenue at each pub; (c) Residuals marked in points and lines are the major roads; (d) Actual against predicted revenue. The experiment resulted in $\Rsqd$ = 0.88 and \NRMSE = 0.03.}
\label{fig:Actual_Pred_vars_edge}
\end{figure}

Next, we perform a detailed study on the model with improved specifications where store features represent the attractiveness of the store (\eq~\eqref{eq:model2}). In this study, the radius of the truncated Gaussian is set to 20km, as it demonstrated the best results for the previous experiment. The model with these settings resulted in a high $\Rsqd$ of 0.88 and a low \NRMSE of 0.03. The plots (\fig~\ref{fig:Actual_Pred_vars_edge}) of the observed revenue and predicted revenues suggest that the model provides a good fit to the data. The residuals are relatively high out of central London, closer to a major ringway as shown in \fig~\ref{fig:Actual_Pred_vars_edge} c. Additionally, the predicted revenue of a few pubs at the edge of the study area is underestimated, reflecting the edge effects. 
 
\begin{figure}[h!]
\centering
	\begin{minipage}{.5\linewidth}
	\centering
	\includegraphics[width=7cm]{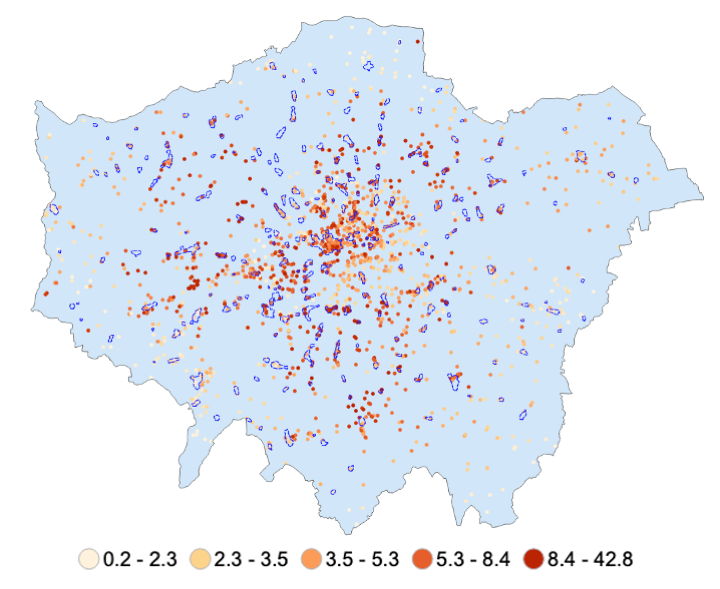}\\
	\scriptsize (a) 
	\end{minipage}%
	\begin{minipage}{.5\linewidth}
	\centering
	\includegraphics[width=7cm]{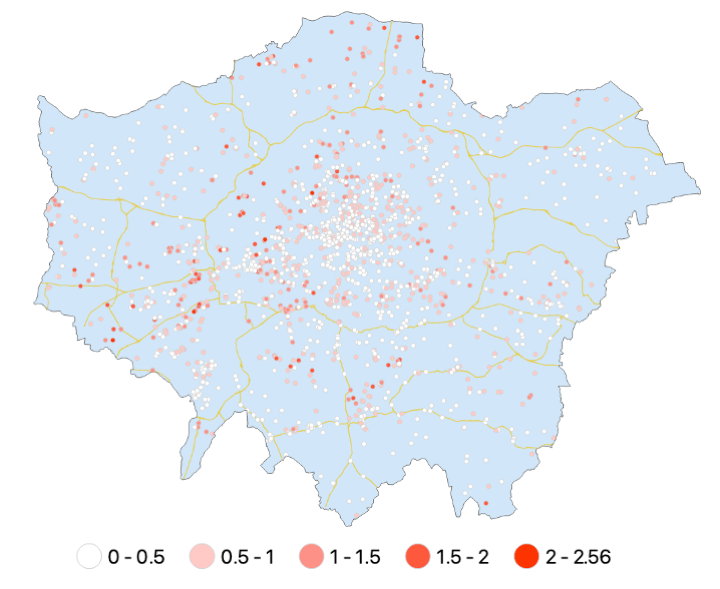}\\ 
	\scriptsize (b) 
	\end{minipage}
\caption{Exploring the pubs attractiveness  for the fitted model: (a) Variance ($\sigma^2_s$) of the Gaussian placed on each pub. Blue colour polygons denote the major towns; (b) Absolute coefficients of the unobserved pub characteristics ($\storeserror_s$). }
\label{fig:store_var}
\end{figure}

Using the parameter estimates ($\veclambda, \vec{\storeserror}$) from the best-fitted model, we have demonstrated the attractiveness ($\sigma^2_s$) of pubs around London in \fig~\ref{fig:store_var}(a). It can be observed that the most attractive pubs are within or around the major towns. Further exploring the coefficients of pub features, we found that Google's customer rating score and the number of people rated had the highest positive contribution towards the attraction term (Appendix C). This implies that customer rating is a critical indicator in describing the customer attractiveness to the pubs. The remaining term used to express the attractiveness, unobserved pub features ($\storeserror_s$), where the absolute coefficient is mapped in \fig~{\ref{fig:store_var}}(b).  There is a similar pattern to the residual plot, but overall spatial distribution appears to be random. A deep investigation is required to understand what could explain the unobserved pub features. 

\begin{figure}[h]
\centering
\includegraphics[width=9cm]{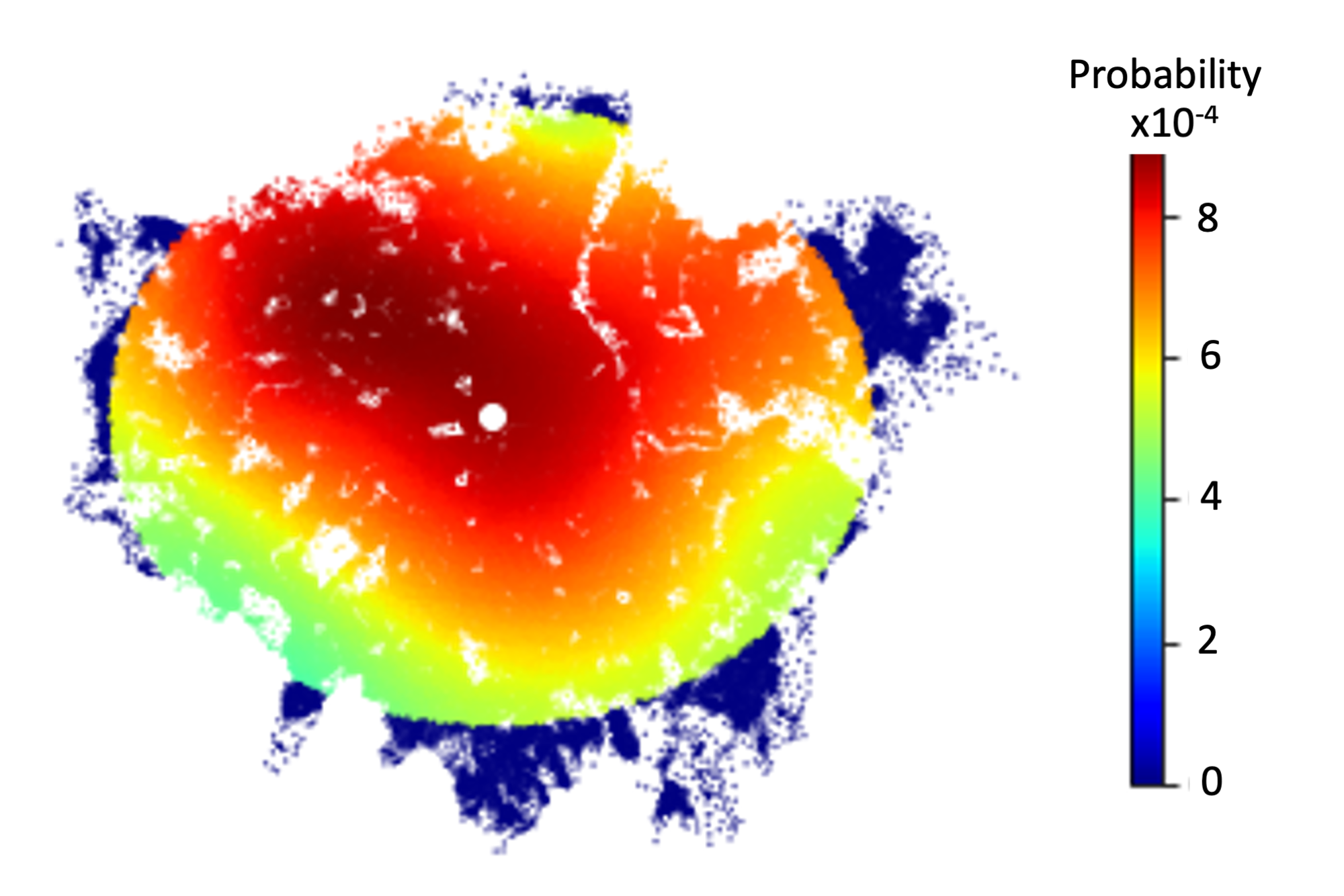}\\ %
\caption{Visualisation of the probability ($\pns$) of people in each postcode selecting the particular pub shown in a white dot in the centre of London.}
\label{fig:pns}
\end{figure}

For demonstration purposes, we randomly select a pub in central London to explore the insights from the fitted model. The probability of people within the postcode selecting the particular pub ($\pns$) is calculated using the model parameter estimates with \eq~\eqref{eq:responsability}. These probabilities are mapped into a heatmap as shown in \fig~\ref{fig:pns}. There appear to be two hotspots on the map, one closer to the pub, and another one towards North-West London. It is natural to see higher probabilities closer to the pub, but the other hotspot is possible because the pubs' density in the area is comparatively low, as shown in \fig~\ref{fig:pubslondon}(a). Hence people in the area also prefer traveling to pubs in central London. The distribution of probabilities tends to be having an oval shape, possibly because the distance between the customers and pubs is calculated as Euclidean distance. A better representation could occur in using a transport network.
\vspace{-0.5cm}

\begin{figure}[h!]
\centering
	\begin{minipage}{.5\linewidth}
	\centering
	\includegraphics[width=6.5cm]{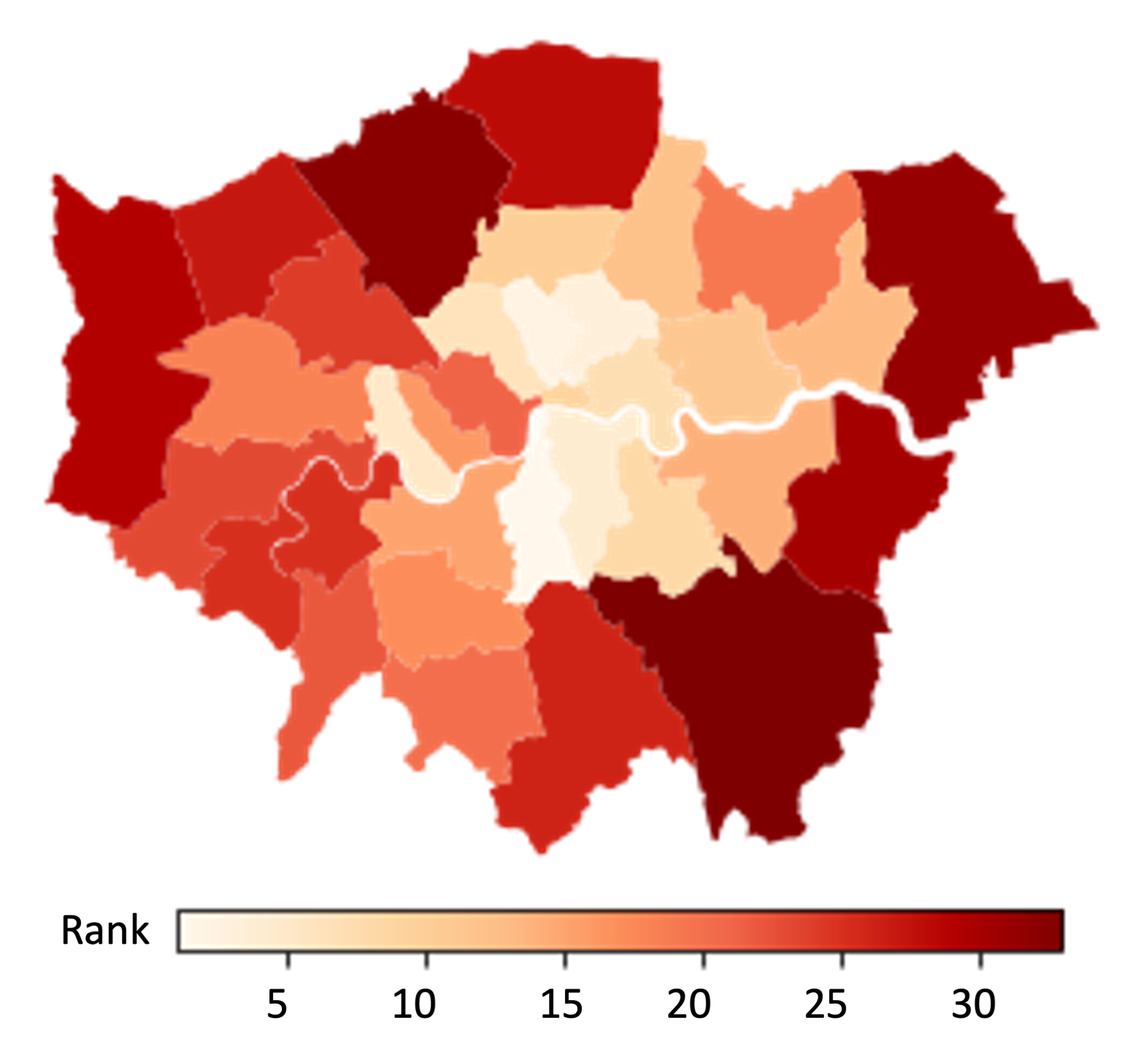}\\
	\scriptsize (a) 
	\end{minipage}%
	\begin{minipage}{.5\linewidth}
	\centering
	\includegraphics[width=6.5cm]{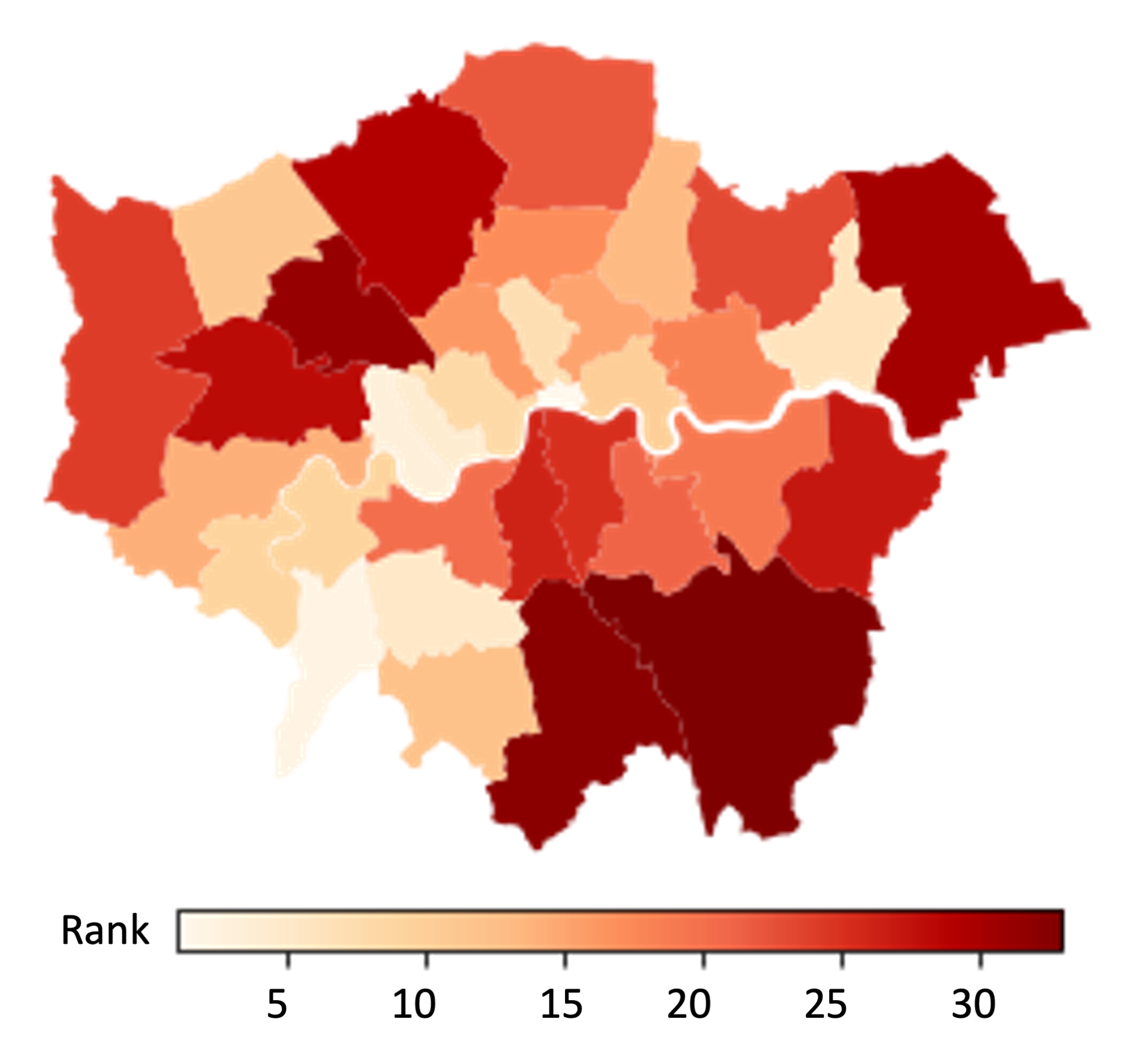}\\ 
	\scriptsize (b) 
	\end{minipage}
\caption{Visualisation of ranking on estimated revenue and mortality in the London Boroughs : (a) Rank of estimated amount spent at the pubs by people living in each Borough; (b) Rank of Mortality count.}
\label{fig:borough_rev}
\end{figure}

Using the parameter estimates ($\vecbeta$) from the best-fitted model, we can estimate the amount spent by customers living in each postcode ($\indrev$). Coefficients of the deprivation features indicate that areas with higher income, high employment, less risk of crimes, better quality of life, and environment tend to positively influence the customers' spending levels at pubs. The amount spent at each Borough can be derived by calculating the total of the estimated spending amount at each postcode within the Borough. This we compare against the alcohol-related mortality in the London Boroughs published by \cite{AlcoholProfile}. The rank of Boroughs respective to the spending and mortality levels published for 2017 is mapped in \fig~\ref{fig:borough_rev}.  The rank correlation between mortality count and estimated spending shows a moderate positive relationship of 0.4. Our intuition is that higher alcohol-related mortalities are to be expected in the areas of high alcohol consumption. 

Finally, we perform a comparison with a spatial interaction model from the literature for completeness of the study. We fitted the Modified Huff model \citep{li2012assessing} for the same dataset, which displayed very low performance with $\Rsqd$ of only 0.03 and \NRMSE of 0.84. Our model outperforms the benchmark model with a notable improvement and provides valuable inferences for decision-makers.

\section{Discussion}	 
We have developed a Bayesian spatial interaction model to simulate customers' behavior with business facilities using their respective characteristics. \our considerably improves existing classical Huff type models as it formally addresses uncertainties arising in the modelling process, via a Bayesian framework while providing inferences at the level of business and customer locations. The key advantage of the proposed model is scalable and can make inference on large-scale datasets through deterministic variational inference, in contrast to the existing models. The synthetic experiments show how \vi performs five times faster than \mcmc while providing comparable performances in terms of parameter identification and without significant under estimation of the posterior covaraince.  

For the first time, we are able to demonstrate and estimate spatial interactions in large real-world urban extend such as Greater London with more than 1500 pubs and 150000 customer regions. For this purpose, we develop a large dataset at the most granular level by utilising data from multiple sources. We presented our methodology in the context of Pubs, but this can be applied in other retail businesses or even expand into sectors such as healthcare and energy. Furthermore, we demonstrate that \our can infer different components of the spatial interactions, thereby making valuable conclusions for a businesses' ability to make decisions. Finally, we have shown how \our outperforms competing approaches in terms of prediction performances while providing consistent results with related indicators observed for the London region.  

The proposed methodology can be extended and improved upon across multiple dimensions. First, one could consider adopting a travel network to estimate the distance instead of the Euclidean metric used in this study \citep{lafferty2005diffusion,grigoryan2009heat,crosby2018road}.  This could provide a more realistic configuration of the geographical setting and lead to better inferences. Furthermore, extending the proposed framework to a spatio-temporal setting to capture the time evolution of parameters to understand the behavioural changes of customers and changes in urban systems will also be of significant interest. Lastly, our work opens up the potential to utilise the \our to select the optimal business location.

\section{Acknowledgements}	
We would like to thank the UK Engineering and Physical Sciences Research Council (EPSRC grant no. EP/L016710/1 and EP/R512229/1). Furthermore, this work was supported by the Alan Turing Institute under EPSRC grant EP/N510129/1, UKRI Turing AI fellowship EP/V02678X/1, and the Lloyds Register Foundation. We are also grateful to Nimbus property system limited for their support and giving access to a comprehensive property database, and for the valuable insights shared by its Director, Paul Davis.
	
\Urlmuskip=0mu plus 1mu\relax	
\bibliographystyle{rss}
\bibliography{example}

\end{document}